\documentclass{article}

\usepackage{microtype}
\usepackage{graphicx}
\usepackage{booktabs} 

\usepackage{hyperref}

\usepackage{fancyhdr}
\usepackage{xcolor} 
\usepackage{algorithm}
\usepackage{algorithmic}
\usepackage{natbib}
\usepackage{eso-pic} 
\usepackage{forloop}
\usepackage{url}

\usepackage[accepted]{icml2024}

\usepackage{amsmath}
\usepackage{amssymb}
\usepackage{mathtools}
\usepackage{amsthm}

\usepackage[capitalize,noabbrev]{cleveref}

\theoremstyle{plain}

\theoremstyle{definition}

\theoremstyle{remark}

\usepackage[textsize=tiny]{todonotes}

\usepackage{amsmath}
\usepackage{listings}
\usepackage{subcaption}
\usepackage{graphicx}
\usepackage{wrapfig}

\DeclareMathOperator*{\argmax}{arg\,max}

\newcommand{\E}{\mathbb{E}}

\newcommand{\secref}[1]{section~\ref{#1}}
\newcommand{\Secref}[1]{Section~\ref{#1}}
\newcommand{\appref}[1]{appendix~\ref{#1}}

\newcommand{\figref}[1]{figure~\ref{#1}}
\newcommand{\Figref}[1]{Figure~\ref{#1}}

\icmltitlerunning{Improving Reward-Conditioned Policies for Multi-Armed Bandits using Normalized Weight Functions}

\begin{document}


\twocolumn[
\icmltitle{Improving Reward-Conditioned Policies for \\
           Multi-Armed Bandits using Normalized Weight Functions}



\icmlsetsymbol{equal}{*}

\begin{icmlauthorlist}
\icmlauthor{Kai Xu}{mitibm}
\icmlauthor{Farid Tajaddodianfar}{amazon}
\icmlauthor{Ben Allison}{amazon}

\end{icmlauthorlist}

\icmlaffiliation{mitibm}{MIT-IBM Watson AI Lab (work done while at Amazon)}
\icmlaffiliation{amazon}{Amazon}

\icmlcorrespondingauthor{Kai Xu}{xuk@mit.edu}

\icmlkeywords{Machine Learning, ICML}

\vskip 0.3in
]



\printAffiliationsAndNotice{} 

\begin{abstract}
Recently proposed reward-conditioned policies (RCPs) offer an appealing alternative in reinforcement learning.
Compared with policy gradient methods, policy learning in RCPs is simpler since it is based on supervised learning, and unlike value-based methods, it does not require optimization in the action space to take actions.
However, for multi-armed bandit (MAB) problems, we find that RCPs are slower to converge and have inferior expected rewards at convergence, compared with classic methods such as the upper confidence bound and Thompson sampling.
In this work, we show that the performance of RCPs can be enhanced by constructing policies through the marginalization of rewards using \emph{normalized weight functions}, whose sum or integral equal $1$, although the function values may be negative.
We refer to this technique as \emph{generalized marginalization}, whose advantage is that \emph{negative} weights for policies conditioned on \emph{low} rewards can make the resulting policies more distinct from them.
Strategies to perform generalized marginalization in MAB with discrete action spaces are studied.
Through simulations, we demonstrate that the proposed technique improves RCPs and makes them competitive with classic methods, showing superior performance on challenging MABs with large action spaces and sparse reward signals.
\end{abstract}

\section{Introduction}
Multi-armed bandit (MAB), or its contextual variant, is a standard formulation for many sequential decision making problems in statistics, engineering, psychology, e-commerce and online advertising \citep{li2010contextual,sutton2018reinforcementlearning}.
Methods for MAB and their reinforcement learning (RL) counterparts are usually value-based or action-based \citep{sutton2018reinforcementlearning}.
In value-based algorithms such as the upper confidence bound (UCB) and Thompson sampling (TS), the value for each action is estimated. 
In order to take an action, it requires to solve an optimization problem in the action space to find the action with the best estimated value.
Such optimization can be expansive or even intractable if the action space is combinatorially large or heterogeneous \citep{hill2017efficient,kumar2019reward}, e.g. when finding the best wiring of eletronic circuits, or optimizing the styles of multiple components in web design.
For policy-based approaches, the most popular approach is gradient bandit, the counterpart of policy gradient in MAB \citep{sutton2018reinforcementlearning}.
For complex problems, the naive policy gradient algorithm is usually difficult and/or expensive to implement and apply \citep{kumar2019reward} because the REINFORCE estimator it uses has high variance.
Thefore some variance reduction techniques with extra baseline models \citep{williams1992simple} are needed, or the doubly robust estimator with extra value estimators \citep{huang2020importancesampling} is required.
This becomes even more challenging when policies are modeled by deep generative models because the large number of parameters often causes training difficulties. 

Reward-conditioned policies \citep[RCPs;][]{kumar2019reward} are a recently proposed alternative to value-based methods and policy gradient methods in RL.
In RCPs, a reward-conditioned policy $\pi_\theta(a \mid r)$ is learned from pairs of reward-action data $(r, a)$ via maximum likelihood estimation of $\theta$, where $a$ is the action, $r$ is the reward and $\theta$ is the parameters of the policy model.
Examples include regression that minimize some error metric \citep{kumar2019reward} or conditional generative modeling that minimize an evidence lower bound \citep[ELBO;][this work]{ajay2023conditionalgenerative}.
After learning, the policy to use at inference time can be constructed through conditioning.
Intuitively, the way how such \emph{inference policies} are constructed can have huge impacts on their performance in terms of convergence speed and expected rewards at convergence.
\citet{kumar2019reward} proposes to use what we refer as the \emph{marginalized policy}, defined as 
$$
\pi_\theta^\dagger(a) = \int \underset{\raisebox{-1em}{\small continuous rewards}}{\pi_\theta(a \mid r) q(r) \mathrm{d}r} \;\text{or}\; \sum_r \underset{\raisebox{-1em}{\small discrete rewards}}{\pi_\theta(a \mid r) q(r)},
$$
where $q(r)$ is the \emph{target value distribution} that models how high reward distributes empirically.\footnote{In Algorithm 1 of \citet{kumar2019reward}, the marginalization is implicitly defined by ancestor sampling, i.e.~first sampling $r'$ from $q(r)$ and then sampling $a$ from $\pi_\theta(a \mid r')$. Note that this marginalization formulation allows the use of a Dirac delta distribution that only has mass on some desirable reward for $q(r)$, which is the actual experiment setup of \citet{kumar2019reward} where some optimistic reward is used (more discussion later).}

In this paper, we apply RCPs to MABs and study how the marginalized policy performs in MAB with discrete actions, which to our best knowledge is not studied in the literature before.
Importantly, we find that the marginalized policy has slower convergence and suboptimal performance than classic methods like the UCB and TS, in both standard and challenging MAB problems.
We therefore propose to use \emph{generalized marginalization} to construct inference policies as\looseness=-1
$$
\pi_{\theta,\phi}^\ddagger(a) = \int \underset{\raisebox{-1em}{\small continuous rewards}}{\pi_\theta(a \mid r) w_\phi(r) \mathrm{d}r} \;\text{or}\; \sum_r \underset{\raisebox{-1em}{\small discrete rewards}}{\pi_\theta(a \mid r) w_\phi(r)},
$$
where $w_\phi(r)$ is a \emph{normalized weight function} with tunable parameter $\phi$, designed to maximize the expected rewards (\secref{sec:method}).
Importantly, $w_\phi(r)$ does not need to satisfy the non-negativity condition but only requires to be \emph{normalized}, i.e.~$\int w_\phi(r) \mathrm{d}r = 1$ (or $\sum_r w_\phi(r) = 1$ for discrete rewards), which is necessary to ensure the normalization property of $\pi_{\theta,\phi}^\ddagger(a)$ as a valid distribution over $a$.
The benefit here is that it allows constructing inference policies that are \emph{dissimilar} to policies conditioned on \emph{low} rewards by having \emph{negative} $w_\phi(r)$ for small $r$.
We study a principled strategy to design and learn $w_\phi$ in settings with discrete actions (\secref{sec:method-gm}) as well as some effective heuristics (\secref{sec:gm-heuristics}).
Through simulations in \secref{sec:sim}, we demonstrate that the proposed technique can outperform the marginalized policy for RCPs in challenging MAB (\secref{sec:sim-challenging}) and contextual MAB (\secref{sec:sim-contextual}) settings, being competitive to classic methods. 
Finally, on MAB with large action spaces and sparse reward signals, we show that RCPs implemented by conditional variational autoencoders (CVAEs) with the proposed techniques can outperform classic methods (\secref{sec:sim-design}).

\section{Related Work}

\paragraph{Multi-armed bandit algorithms}
The UCB and TS algorithms are principled methods for MAB with provable convergence bounds and are widely used in industries.
\citet{li2010contextual} proposes to solve the contextual bandit problem using linear models in UCB.
\citet{hill2017efficient} uses TS in which the optimization step is solved by hill-climbing to optimize web layout design, for which the action space is combinatorially large.
UCB with neural networks are explored by many works with different ways to model the confidence level, via first-order approximation \citet{zhou2020neural} or using a separate network to model the variance \citep{zhu2021deepbandits}.
Similarly, TS with neural networks are studied with different ways for approximate Bayesian inference, via Gaussian approximations \citep{zhang2020neural} or sample average uncertainty \citep{rawson2021deepupper}.
Standard RL algorithms have also been adopted to solve bandit problems.
Policy gradient has been used to solve bandit problems for contextual recommendation \citep{pan2019policy}.
\citet{cheng2022policy} uses policy gradient together with a generative model for chip design optimization.
Actor-critic algorithms \citet{konda1999actorcriticalgorithms} have also been used for bandit problems \citep{yu2018reducingvariance,lei2022actorcriticcontextual}.

\paragraph{Reward-conditioned policies}
\citet{kumar2019reward} propose to learn a reward-conditioned policy as a regression task with concurrent work referring this approach as upside-down reinforcement learning \citep{srivastava2019trainingagents,schmidhuber2019reinforcementlearning}.
\citet{ding2023bayesian} uses Bayesian reparameterization to improve generalization on high reward-to-go (RTG) inputs and to avoid out-of-distribution RTG queries during testing time.
Similarly, goal-conditioned RL requires learning policies according to different goals \citep{liu2022goal,li2022phasic,wang2023optimal}.
Reward-conditioned policies has also been studied under the name reinforcement learning via supervised learning \citep[RvS;][]{emmons2022rvswhat} or reward-conditioned supervised learning \citep[RCSL;][]{brandfonbrener2023whendoes}.
These works find that RCPs enjoys better performance than other offline RL methods but require a stronger assumption than traditional dynamic programming based methods.

\paragraph{Conditional generation for optimization and decision-making}
Conditional generative models have been explored in many recent works to solve optimization and decision-making problems.
\citet{jiang2019greedyranking} uses CVAEs for slate optimization problems by learning the joint distribution of components on the slate conditioned on user responses.
\citet{chen2021decisiontransformer} propose to formulate decision-making as sequence modeling tasks where the state, action and reward at each step are consumed by a transformer to output the action, which can be also seen as an instantiation of reward-conditoned policies.
Follow-up works generalize this idea to other models such as diffusion models \citep{ajay2023conditionalgenerative} or with more complex setups with online components \citep{zheng2022onlinedecision}.
For generation tasks, \citet{kanungo2022cobartcontrolleda} explores conditioning on positive click-through-rates to improve headline generation by fintuning pretrained language models for reward conditioning.

\section{Generalized Marginalization for Reward-Conditioned Policies}\label{sec:method}
In this section,
we start by introducing the general formulation of generalized marginalization of reward-conditioned policies for MAB with discrete action spaces (\secref{sec:method-gm}), where we start with binary rewards and extend it to other reward types in \secref{sec:other-reward-types}. We then discuss two simple and computationally cheap heuristic strategies for it (\secref{sec:gm-heuristics}).
We finish by illustrating how the resulting inference polices differ from each other and giving inuitions of why they are expected to perform better (\secref{sec:illustration}).

\subsection{Generalized marginalization with discrete action spaces}\label{sec:method-gm}
We start our discussion with RCPs for contextual MAB with binary rewards.
Assume that we have obtained some reward-conditoned policy $\pi_\theta(a \mid c, r)$ where $a$ is the action, $c$ is the context and $r$ is the reward. 
Policy inference by generalized marginalization with a general normalized weight functions
$$
w_\phi(r)=\begin{cases}w_0 & r=0 \\w_1 & r=1\end{cases}\quad\text{subject to}\quad w_0 + w_1 = 1
$$
gives inference policies $\pi_{\theta,\phi}^\ddagger(a \mid c)$ with the following form
\begin{equation}\label{eq:gm-inference}
  \begin{aligned}
    \pi_{\theta,\phi}^\ddagger(a \mid c)
    &= \sum_{r'=0,1} \pi_\theta(a \mid c, r=r') w_\phi(r=r') \\
    &= \pi_\theta(a \mid c, r=0) w_0 + \pi_\theta(a \mid c, r=1) w_1
  \end{aligned}
\end{equation}
where the weight parameter $\phi$ is a 2-dimensional vector $[w_0, w_1]$.
Also note that $\phi$ has a different parameratization that always satisfies the noramlization constraint: $\phi=[1 - \lambda, \lambda]$, which we will use later.
Our goal is to optimize $\phi$ such that the expected reward is maximized
\begin{equation}\label{eq:maximizing-expected-reward}
  \phi^\ast = \argmax_\phi \E_{p_C} \E_{\pi_{\theta,\phi}^\ddagger(a \mid c)} [\bar{R}(a \mid c)]
\end{equation}
where $p_C$ is the context distribution, $\bar{R}(a \mid c)$ is the unknown value function of action $a$ under context $c$.
As $w_0,w_1$ do not necessarily have to be positive, it is now possible to have $w_0<0$ and $w_1>0$ to construct policies that is dissimilar to $\pi_\theta(a \mid c, r=0)$, 
which intuitively could obtain a larger expected reward; we will also refer to $\pi_\theta(a \mid c, r=0)$ as the \emph{negative policy}, and similarly $\pi_\theta(a \mid c, r=1)$ as the \emph{positive policy}, in the rest of the paper.
Note that, although the formulation ensures that the inference policy satisfies the normalization property as a valid probability distribution, 
the non-negativity still needs to be ensured, i.e.
\begin{equation}\label{eq:valid-policy-condition}
\pi_{\theta,\phi}^\ddagger(a \mid c) \geq 0 \quad\forall\quad a.
\end{equation} 
We will show how it constrains the range of $\lambda$ (the only variable in $w_\phi$) for discrete action spaces next.

Substituting \eqref{eq:gm-inference} into \eqref{eq:maximizing-expected-reward} with the parameratization $w_0 = 1- \lambda, w_1 = \lambda$, we have the optimization task
\begin{equation}\label{eq:objective}
  \begin{aligned}
    &\argmax_{\lambda} R(\lambda)\quad\text{where}\\
    &\begin{aligned}
      R(\lambda) := \E_{p_C} \{ &\E_{\pi_{\theta}(a \mid c, r=0)} [(1 - \lambda) \bar{R}(a \mid c)] + \\
                                &\E_{\pi_{\theta}(a \mid c, r=1)} [\lambda \bar{R}(a \mid c)] \}.
    \end{aligned}
  \end{aligned}
\end{equation}
Given $N$ off-policy data points, $\{(c^i, a^i, r^i, q^i)\}_{i=1}^N$, where $r^i$ is the observed reward and $q^i := q(a^i \mid c^i)$ is the action probability of the data collection policy, $R(\lambda)$ can be estimated using Monte Carlo with importance sampling as
\begin{equation}\label{eq:objective-is}
  \begin{aligned}
    R(\lambda) 
    &\approx \frac{1}{N} \sum_{i=1}^N \{{\pi_{\theta}(a \mid c, r=0) \over q^i} (1 - \lambda) r^i + \\
    &\qquad\qquad\qquad{\pi_{\theta}(a \mid c, r=1) \over q^i} \lambda r^i \} \\
    &= \frac{\lambda}{N} 
    \{ \sum_{i=1}^N \left[ {\pi_{\theta}(a \mid c, r=1) \over q^i}r^i \right] - \\
    &\qquad\qquad\sum_{i=1}^N \left[ {\pi_{\theta}(a \mid c, r=0) \over q^i} r^i \right] \}   + \text{constant}
  \end{aligned},
\end{equation}
which is linear to $\lambda$.
In other words, depending on the sign of the difference term (the coefficient of $\lambda$), one either needs to use the maximum or minimal of $\lambda$ under the constraint specified by \eqref{eq:valid-policy-condition}, which can be solved efficiently as below.
For discrete action space with $K$ actions $a_1, \dots, a_K$ and for a given context $c$, \eqref{eq:valid-policy-condition} reduces to a set of constraints on $\lambda$ as:
$$
  (1 - \lambda) \pi_0^i + \lambda \pi_1^i \geq 0 \quad\text{for}\quad i=1, \dots, K,
$$
where we denote $\pi_0^i = \pi_\theta(a_i \mid c, r=0)$ and $\pi_1^i = \pi_\theta(a_i \mid c, r=1)$ to simplify the notation.
This gives lower and upper bounds for $\lambda$ as:
\begin{equation}\label{eq:bounds}
  \begin{aligned}
    &\max(\Lambda_l) \leq \lambda \leq \min(\Lambda_u)\quad\text{where}\\
    &\begin{aligned}
      \Lambda_u &:= \{-{\pi_0^i \over \pi_1^i - \pi_0^i} \mid \pi_1^i - \pi_0^i > 0\} \quad\text{and}\\
      \Lambda_l &:= \{-{\pi_0^i \over \pi_1^i - \pi_0^i} \mid \pi_1^i - \pi_0^i < 0\}
    \end{aligned}
  \end{aligned}.
\end{equation}

Therefore, solving \eqref{eq:objective} is reduced to two steps: (i) computing the upper and the lower bounds according to \eqref{eq:bounds} and (ii) computing \eqref{eq:objective-is} at two bounds and picking $\lambda$ that yields the larger $R(\lambda)$ estimate.
We refer an inference policy resulting from this optimization procedure of our generalized marginalization strategy as the \emph{optimized policy}.

\subsubsection{Beyond binary rewards}\label{sec:other-reward-types}
We now discuss ways to extend our proposed method beyond binary rewards.
For discrete rewards with finite values, instead of $0$ and $1$, it is straightforward to set the conditioning rewards for the negative and positive policy to the minimal and maximal values in the discrete space, respectively.
For continuous rewards, we propose to set the conditioning rewards to the $q_0$-percent and $q_1$-percent quantiles, e.g. $q_0=10, q_0=90$, from observed rewards, where $q_0,q_1$ are hyperparameters to set for low and high percentiles; \citet{kumar2019reward} uses the same idea to select the optimistic reward.
For discrete rewards with unbounded values, a similar strategy can be used with the quantile values cast to their closest discrete values in the domain.\footnote{Note that these ways to deal with general discrete rewards by transforming into a form of low and high rewards are designed such that the effcient solution developed in \secref{sec:method-gm} can be used.
Other ways under the idea of generalized marginalization that require solving a more complex optimization problem are mentioned in \secref{sec:discussion}.}

\subsection{Heuristic strategies for generalized marginalization}\label{sec:gm-heuristics}
Computing \eqref{eq:objective-is} for each context $c$ can still be computationally expansive for large action spaces, therefore we further propose two heuristics that share the same intuition as the solution developed in \secref{sec:method-gm}.

\paragraph{Optimistic policy}
An intuitive strategy to approach \eqref{eq:maximizing-expected-reward} would be to set $w_0=0, w_1=1$, which satisfies all constraints by design.
This is essentially using the positive policy (i.e.~the reward-conditioned policy conditioned on the positive reward) and is similar to what \citet{kumar2019reward} do in their experiments, despite that their settings is real-valued rewards where they use empirically large rewards observed in the data as the reward condition.
As the negative policy is not used at all, we do not expect it to work well, which is later confirmed in our simulation.

\paragraph{SubMax policy}
With our original intuition being to make policies dissimilar to the negative policy $\pi_\theta(a \mid c, r=0)$,
here we present another simple heuristics to maximize \eqref{eq:maximizing-expected-reward} for discrete actions.
Denoting $\mathbf{p}_0 := \pi_\theta(a \mid c, r=0), \mathbf{p}_1 := \pi_\theta(a \mid c, r=1)$, we define the \emph{SubMax policy} as
\begin{equation}\label{eq:gm-inference-submax}
  \pi_{\theta,\phi}^\ddagger(a \mid c) = \mathrm{normalize}\left(\max(\mathbf{p}_1 - \mathbf{p}_0, 0)\right),
\end{equation}
where $\max$ is applied to the vector in an element-wise manner and $\mathrm{normalize}$ is the operator that normalizes a vector to have a sum of $1$.
The name SubMax refers to the operations we take in \eqref{eq:gm-inference-submax}: the subtraction of the two probability vectors followed by the element-wise maximization against $0$.
This SubMax strategy can in fact lead to polices that are outside the space defined by \eqref{eq:gm-inference}, but such construction follows the same motivations which is to be dissimilar to the negative policy. 
To see this, note that \eqref{eq:gm-inference-submax} could lead to an inference policy with $0$ probabilities for multiple actions while \eqref{eq:objective} would only lead to an policy with $0$ probability for a single action, unless more than one actions attain equality in \eqref{eq:valid-policy-condition} for the same $\lambda$.
The illustration in \secref{sec:illustration} visually shows this.
Unexpectedly, we find the SubMax policy to be generally competitive or even superior than other methods in our simulation.

\subsection{Illustration of inference polices}\label{sec:illustration}

\begin{figure*}[ht]
  \centering
  \begin{subfigure}[b]{0.275\textwidth}
      \centering
      \includegraphics[width=\textwidth]{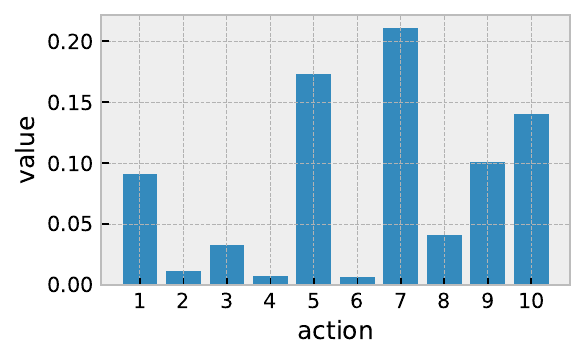}
      \caption{environment (0.0816)}
      \label{fig:illustration-env}
  \end{subfigure}
  \begin{subfigure}[b]{0.275\textwidth}
      \centering
      \includegraphics[width=\textwidth]{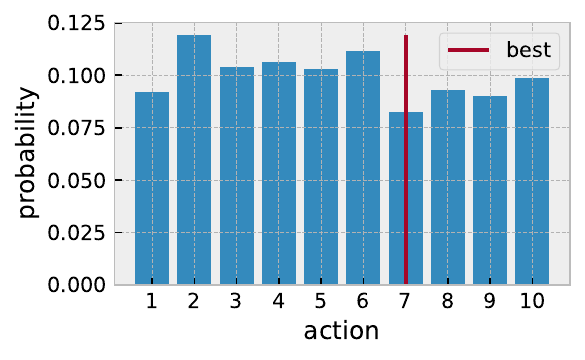}
      \caption{$0$-conditioned policy ($0.077$)}
      \label{fig:illustration-r=0}
  \end{subfigure}
  \begin{subfigure}[b]{0.275\textwidth}
      \centering
      \includegraphics[width=\textwidth]{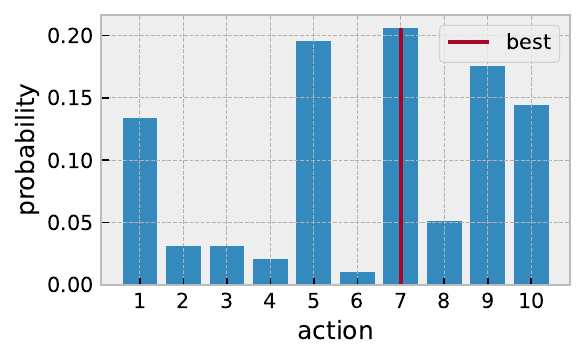}
      \caption{$1$-conditioned policy ($0.131$)}
      \label{fig:illustration-r=1}
  \end{subfigure}
  \\
  \begin{subfigure}[b]{0.275\textwidth}
      \centering
      \includegraphics[width=\textwidth]{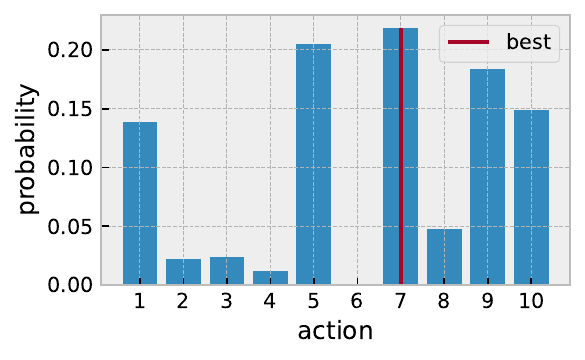}
      \caption{optimized policy ($0.136$)}
      \label{fig:illustration-optimal}
  \end{subfigure}
  \begin{subfigure}[b]{0.275\textwidth}
      \centering
      \includegraphics[width=\textwidth]{./figs/illustration-optimistic.pdf}
      \caption{optimistic policy ($0.131$)}
      \label{fig:illustration-optimistic}
  \end{subfigure}
  \begin{subfigure}[b]{0.275\textwidth}
      \centering
      \includegraphics[width=\textwidth]{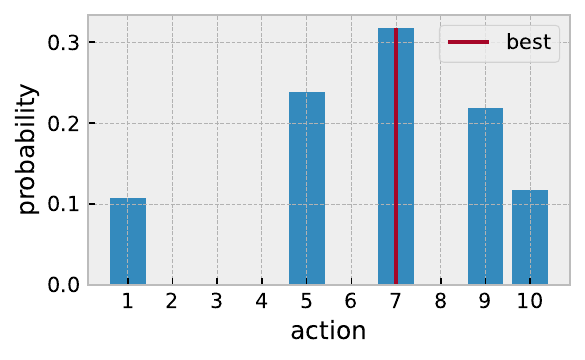}
      \caption{SubMax policy ($0.157$)}
      \label{fig:illustration-SubMax}
  \end{subfigure}
     \caption{Illustration of different strategies to construct inference policies. \Figref{fig:illustration-env} shows the true value per action and the rest of the figures show different inference-time policies where the red line indicates the ground truth best arm. Numbers in braces are the expected reward (larger the better) under each policy and the number next to the environment corresponds to a random policy. Note that the 1-conditioned policy, or positive policy, is equivalent to the optimistic policy.}
     \label{fig:illustration}
\end{figure*}

We now visualize the inference polices from different strategies in a simple, non-contextual MAB.
We consider a problem with $10$ Bernoulli arms with probabilities sampled from $\mathrm{Beta}(1, 9)$ and collect $1000$ observations using a random policy (i.e. uniformly randomly picking $10$ arms). 
The probabilities in the reward condition policy, $\pi_\theta(a \mid r=0)$ and $\pi_\theta(a \mid r=1)$, can be estimated by simply counting the action frequencies for with $r=0$ and $r=1$, respectively.
We then apply different strategies to construct inference policies using generalized marginalization: optimized policy, optimistic policy and SubMax policy.
\Figref{fig:illustration} shows the ground truth value (i.e. expected reward) for each action in the environment, the two reward-conditioned polices and the resulting inference policies from three strategies.
As it can be seen, all policies except the $0$-reward policy or negative policy ($0.077$) achieves better than random expected reward ($0.0816$).
Importantly, the optimistic policy, equivalently the $1$-conditioned policy, which does not use the negative policy, is the worst ($0.131$) among the rest.
Optimized policy ($0.136$) perform better than the optimistic one as expected, since it directly maximizes the expected reward \eqref{eq:objective}.
Unexpectedly, the SubMax policy ($0.157$) works the best, despite of its simplicity.
Qualitatively, we can see that both optimized policy and SubMax policy have the ability to ``zero out'' certain actions that have bad performance (estimated value) so far.
However, due to the set of constraints in \eqref{eq:bounds} that the optimized policy has to satisfy, in general it can only ``zero out'' one action, unless more than one actions attain equality in \eqref{eq:valid-policy-condition} for the same $\lambda$, while the SubMax operation can potentially ``zero out'' more through the $\max$ operator.
This is shown in \figref{fig:illustration-optimal} where only action 6 has a probability of $0$ while in \figref{fig:illustration-SubMax}, action 2, 3, 4, 6 and 8 all have $0$ probabilities.

\section{Simulations}\label{sec:sim}
In this section, we perform a set of MAB simulations with diverse settings.
\Secref{sec:sim-challenging} focuses on simple non-contextual MAB with sparse rewards, large action space and delayed reward, respectively, where we also compare the proposed method with classic MAB algorithms.
\Secref{sec:sim-contextual} further studies the method in contextual MAB and compare it against TS.
Finally, we design a contextual MAB problem with large combinatorial action spaces in \secref{sec:sim-design} to study how the proposed method works for design problems where multiple discrete design choices need to be made jointly.

We implement algorithms studied in this section as a Julia package called \texttt{MultiArmedBandits.jl} for reproducibility; codes can be found in the supplementary material.

\subsection{RCPs and classic methods on challenging non-contextual MABs}\label{sec:sim-challenging}

For simulation of non-contextual MABs, similarly to \secref{sec:illustration},
we randomly initialize $K$ Bernoulli arms with a predefined beta distribution $\mathrm{Beta}(\alpha, \beta)$.
To mimic real-world reward delay, we implement a delay buffer that only returns the simulated data after $N_b$ observations.
We focus on studying how varying $K$, $(\alpha, \beta)$ and $N_b$ affects the convergence of each algorithm, as indicated by the accumulated regret over time.
The sweeping settings we use are (1) $K = 10, 100, 500$, (2) $(\alpha, \beta) = (1, 1), (1, 9), (1, 99)$ and (3) $N_b = 100, 500, 1000$, which results in 27 total configurations.
For each configuration, we randomly initialize an environment and run each algorithm for $5000$ observations; we repeat such simulation for $100$ times and plot the mean along with its 95\% quantiles in ribbons.

For comparison, we consider the following list of MAB algorithms: (1) random: a policy with uniformly random selection, (2) $\epsilon$-greedy: a greedy policy with a chance of $\epsilon$ for picking the arm with the best estimated value and otherwise uniformly randomly selecting the arm, (3) UCB1: upper confidence bound, (4) TS: Thompson sampling with independent beta-Bernoulli arms, (5) RCPs with optimistic policies, (6) RCP: RCPs with the optimized policy and (7) the SubMax policy.
As the choice of $\epsilon$ in (2) and the choice of prior in (4) highly affect their performance, we further sweep these hyperparameters.

Here, due to space limitation, we focus on discussing the results of varying one parameter while keeping the other two fixed; the results for all 27 configurations is in \appref{app:exp-challenging}\footnote{Our appendix is provided as a separate file, \texttt{appendix.pdf}, in the supplementary material due to over-sized PDF.}.

\begin{figure*}[ht]
  \centering
  \begin{subfigure}[b]{0.32\textwidth}
      \centering
      \includegraphics[width=\textwidth]{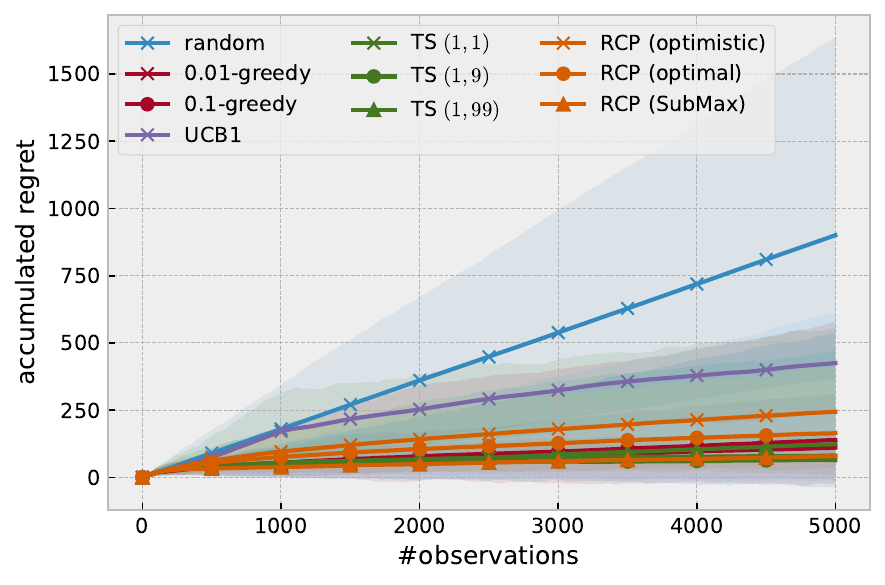}
      \caption{$K=10$}
      \label{fig:exp-1-K=10}
  \end{subfigure}
  \hfill
  \begin{subfigure}[b]{0.32\textwidth}
      \centering
      \includegraphics[width=\textwidth]{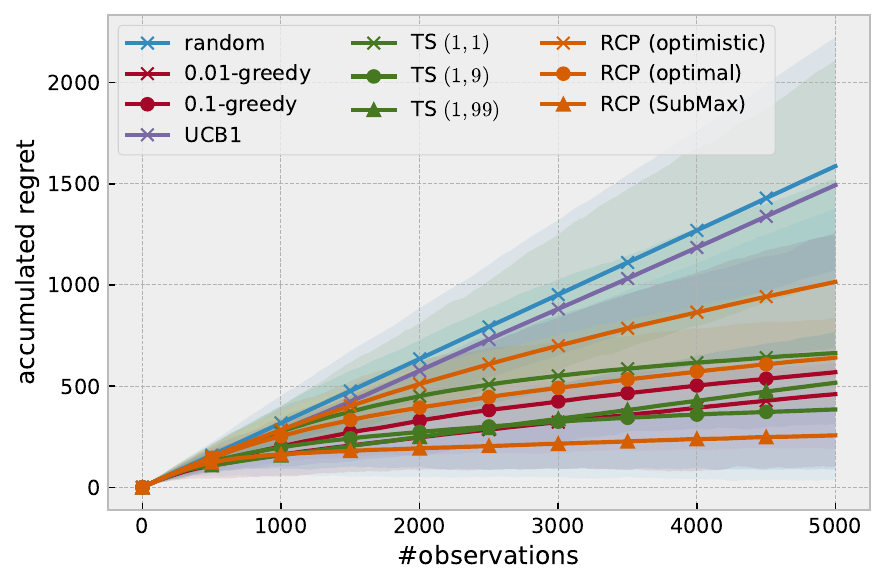}
      \caption{$K=100$}
      \label{fig:exp-1-K=100}
  \end{subfigure}
  \hfill
  \begin{subfigure}[b]{0.32\textwidth}
      \centering
      \includegraphics[width=\textwidth]{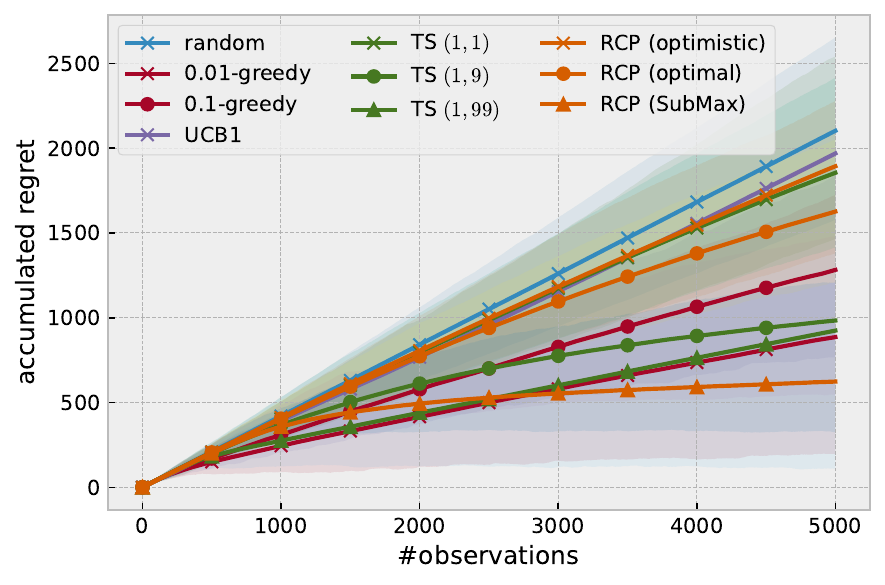}
      \caption{$K=500$}
      \label{fig:exp-1-K=500}
  \end{subfigure}
     \caption{Accumulated regret with varying $K$ and $(\alpha, \beta) = (1, 9)$ and $N_b = 500$.}
     \label{fig:exp-1-K}
\end{figure*}

\paragraph{Large action space}
We first look at the results of varying $K$ while keeping $(\alpha, \beta) = (1, 9), N_b = 100$, which is shown in \figref{fig:exp-1-K}.
We see that the performance of classic methods degrade when $K$ increase, except TS with the prior equal to the ground truth beta used for initializing the environment.
Overall, RCPs with the SubMax strategy yields the best policy across the 3 settings. 
For the rest two strategies, the optimized strategy is consistently better than the optimistic one.\looseness=-1

\begin{figure*}[ht]
  \centering
  \begin{subfigure}[b]{0.32\textwidth}
      \centering
      \includegraphics[width=\textwidth]{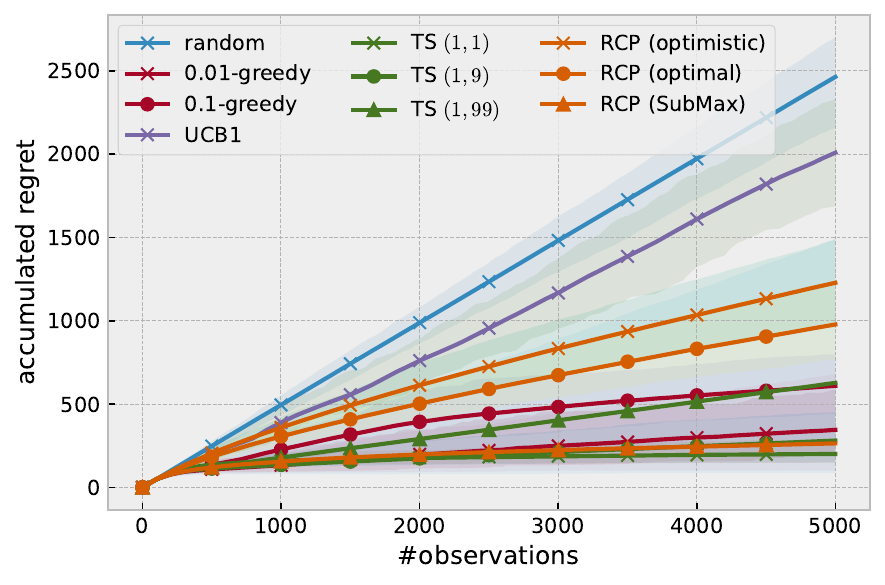}
      \caption{$\mathrm{Beta}(1, 1)$}
      \label{fig:exp-1-ab=uniform}
  \end{subfigure}
  \hfill
  \begin{subfigure}[b]{0.32\textwidth}
      \centering
      \includegraphics[width=\textwidth]{./figs/exp-1/K=100-Nb=100-ab=sparse.pdf}
      \caption{$\mathrm{Beta}(1, 9)$}
      \label{fig:exp-1-ab=sparse}
  \end{subfigure}
  \hfill
  \begin{subfigure}[b]{0.32\textwidth}
      \centering
      \includegraphics[width=\textwidth]{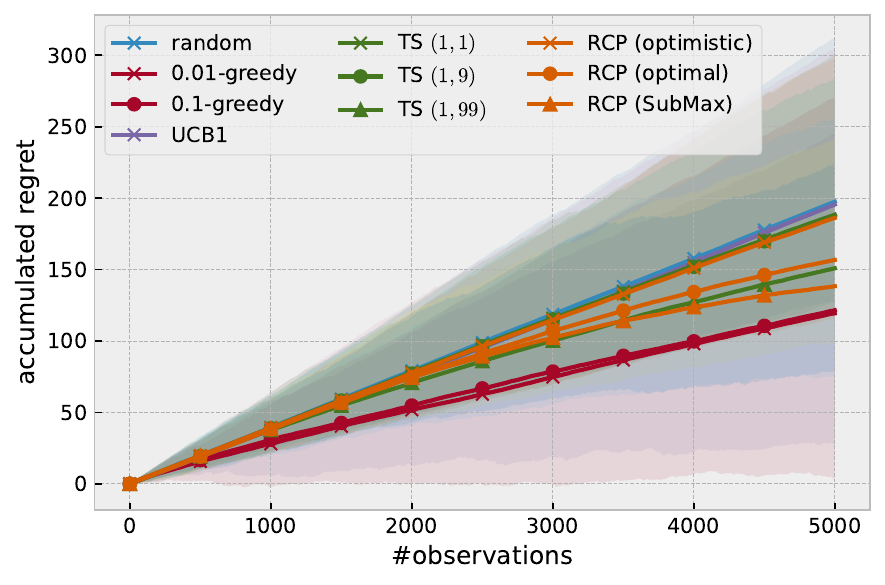}
      \caption{$\mathrm{Beta}(1, 99)$}
      \label{fig:exp-1-ab=xsparse}
  \end{subfigure}
     \caption{Accumulated regret with varying $(\alpha, \beta)$ while keeping $K = 100$ and $N_b = 100$.}
     \label{fig:exp-1-ab}
\end{figure*}

\paragraph{Sparse reward}
We then look the results of varying $(\alpha, \beta)$ while keeping $K = 100, N_b = 100$ in \figref{fig:exp-1-ab}.
Here RCPs with the SubMax strategy still remains competitive across different methods.
For each prior, TS with the prior being the same one used to initialize the environment is also better than others and for $\mathrm{Beta}(1, 99)$, $\epsilon$-greedy strategy show a faster convergence within $5000$ observations.

\begin{figure*}[ht]
  \centering
  \begin{subfigure}[b]{0.32\textwidth}
      \centering
      \includegraphics[width=\textwidth]{./figs/exp-1/K=100-Nb=100-ab=sparse.pdf}
      \caption{$N_b=100$}
      \label{fig:exp-1-Nb=100}
  \end{subfigure}
  \hfill
  \begin{subfigure}[b]{0.32\textwidth}
      \centering
      \includegraphics[width=\textwidth]{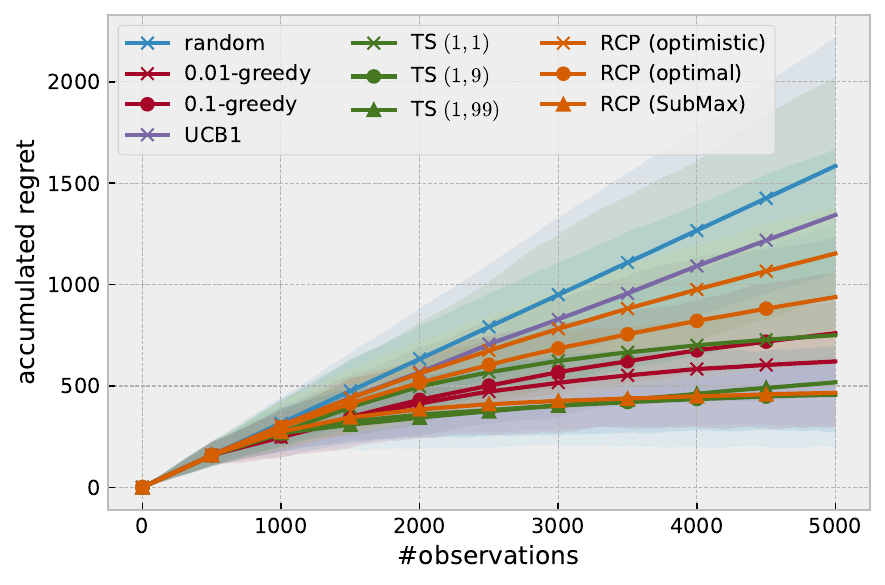}
      \caption{$N_b=500$}
      \label{fig:exp-1-Nb=500}
  \end{subfigure}
  \hfill
  \begin{subfigure}[b]{0.32\textwidth}
      \centering
      \includegraphics[width=\textwidth]{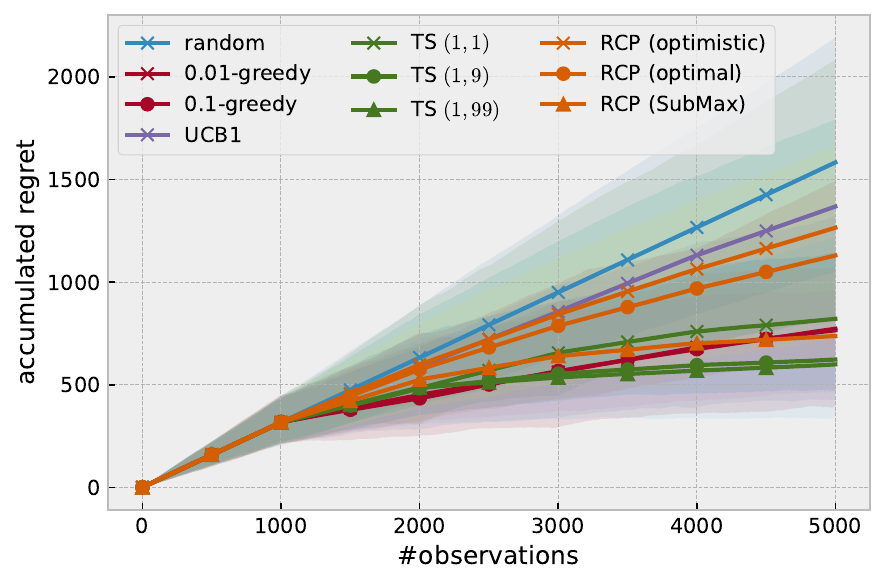}
      \caption{$N_b=1000$} 
      \label{fig:exp-1-Nb=1000}
  \end{subfigure}
     \caption{Accumulated regret with varying $N_b$ while keeping $K = 100$ and $(\alpha, \beta) = (1, 9)$.} 
     \label{fig:exp-1-Nb}
\end{figure*}

\paragraph{Delayed reward}
Finally, we check the results of varying $N_b$ while keeping $K = 100, (\alpha, \beta) = (1, 9)$, which is shown in \figref{fig:exp-1-Nb}.
Again RCPs with the SubMax strategy still remains competitive across different methods.
It is only outperformed by TS with the largest reward delay. 
However, based on the trend (the slope of the accumulated regret curve), RCPs with the SubMax policy is likely to outperform TS beyond the $5000$ observations.

Overall, RCPs with the SubMax strategy is the most competitive method across all configurations and often converge faster than classic methods especially when the MAB settings are difficult.
Within RCPs, the SubMax strategy outperforms the optimized strategy which again outperforms the optimistic strategy.\looseness=-1

\subsection{RCPs and TS on contextual MABs}\label{sec:sim-contextual}
To perform simulations with contextual MABs, we follow the setup in \citet{zhou2020neural}.
For each environment, we randomly initialize a neural network to represent ground truth value distribution: it takes in a pair of action and context and returns the parameter of a Bernoulli distribution.
During simulation the context is sampled from a hypersphere uniformly.
We use $K=4$ actions and contexts with $20$ dimensions, which is the same as \citet{zhou2020neural}.
Additionally, we use a delay buffer of size $N_b=1000$ and also add a constant shift $s$ to the output of the value function in the logit space in order to control the sparsity of the reward (a negative shift would lead to a Bernoulli with small chances of positive rewards).
We vary $s = 0, -2.0, -4.0$ and for each configuration, we randomly initialize an environment and run the algorithm for $10000$ observations; we repeat such simulation for $50$ times and plot the mean of accumulated regret with its 95\% quantiles in ribbons.

We consider comparing two RCP variants studied in the paper against two classic MAB lagorithms, $\epsilon$-greedy and TS; we do not study RCPs with the optimized strategy here because it is computationally expansive to compute \eqref{eq:objective-is} per context, which is needed for contextual MABs.
In RCPs, the policies are modeled by CVAEs \citep{kingma2014semisupervisedlearning} with multiplicative injection layers \citep{kumar2019reward}.\footnote{Using a CVAE is not necessarily for this experiment as the output is a univariate Categorical. The intention here is to make the implementation consistent with that in \secref{sec:sim-design} where the policy requires a deep generative model to capture correlations in the action space. We detail the use of CVAEs in the next section.}
For TS, we use neural networks to fit the value function and use Monte Carlo dropout to approximate model uncertainty, i.e.~to sample a random value we run a forward pass of the network with dropout enabled.
Finally, we sweep $\epsilon$ for $\epsilon$-greedy as it affects its performance.

\begin{figure*}[t]
  \centering
  \begin{subfigure}[b]{0.32\textwidth}
      \centering
      \includegraphics[width=\textwidth]{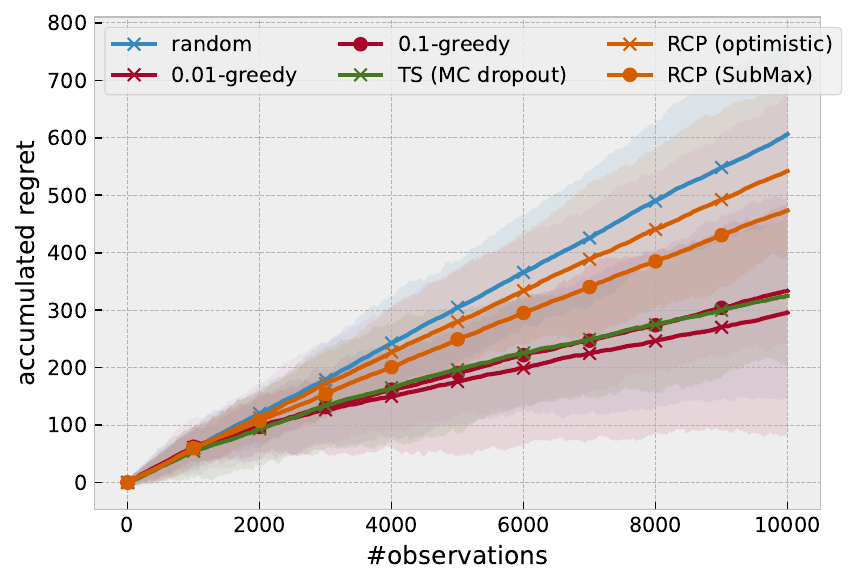}
      \caption{$s=0$ ($\E[\bar{R}] = 0.501$)}
      \label{fig:exp-2-uniform}
  \end{subfigure}
  \hfill
  \begin{subfigure}[b]{0.32\textwidth}
      \centering
      \includegraphics[width=\textwidth]{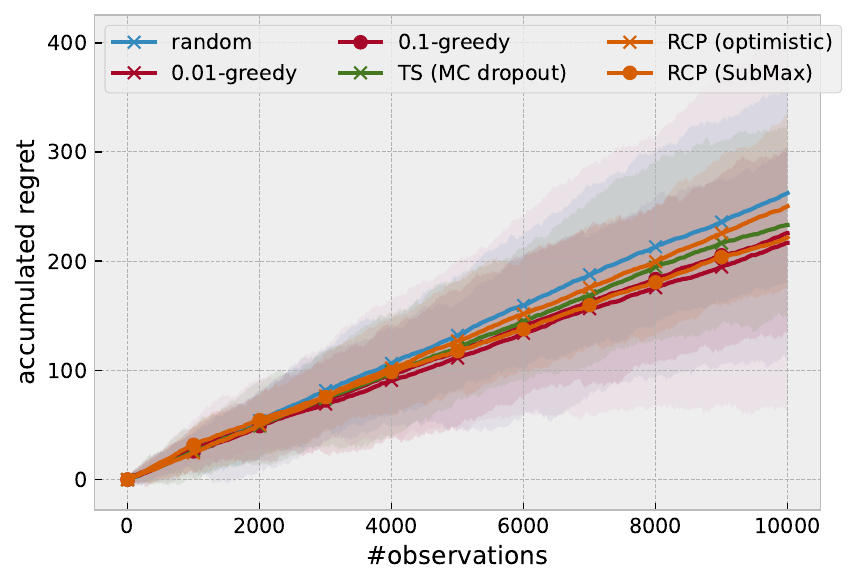}
      \caption{$s=-2.0$ ($\E[\bar{R}] = 0.123$)}
      \label{fig:exp-2-sparse}
  \end{subfigure}
  \hfill
  \begin{subfigure}[b]{0.32\textwidth}
      \centering
      \includegraphics[width=\textwidth]{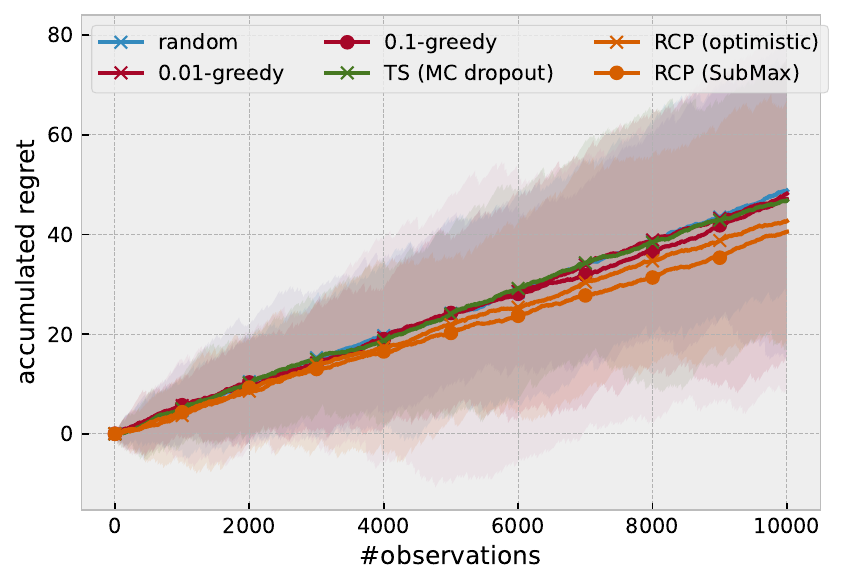}
      \caption{$s=-4.0$ ($\E[\bar{R}] = 0.0185$)}
      \label{fig:exp-2-xsparse}
  \end{subfigure}
     \caption{Accumulated regret with varying constant shift $s$ in the ground truth value function. The numbers next to each $s$ in the subtitles are the average of the expected reward over all randomized environment.}
     \label{fig:exp-2}
\end{figure*}

The accumulated regret plots for this simulation setup are given in \Figref{fig:exp-2}.
As we can see, while classic methods perform better in a dense reward setting, their performance degrade a lot when the rewards become sparse.
RCPs, on the other side, are more robust to the sparsity in reward, rendering RCPs with SubMax being the best in the highest sparsity setup (\figref{fig:exp-2-xsparse}).
One explanation for the success of RCPs with the SubMax policy is that, the $0$-conditioned policy, which is used to construct the inference policy, benefits from the increasing number of negative rewards during training.

\subsection{RCPs on MAB with combinatorial action spaces}\label{sec:sim-design}
Finally, we study how RCPs with different polices perform in MAB with large combinatorial action spaces.
In particular, we define the action to be the combination of 5 options $a = [a_1, \dots, a_5]$ chosen from a set of $2,4,6,16,32$ possible choices, resulting an action space with cardinality of $2\times4\times6\times16\times32 = 32768$ in total.
The rest of environment setup is same as that of \secref{sec:sim-contextual} where the one-hot choices are concatenated as the action to be consumed by the ground truth value function.
We randomly initialize an environment and run the algorithm for $10000$ observations; we repeat such simulation for $20$ times and plot the mean of the accumulated \emph{reward} along with 95\% quantiles.
Note that we plot the reward instead of regret because it is expensive to calculate the optimal action/value required by regret due to the large action space.

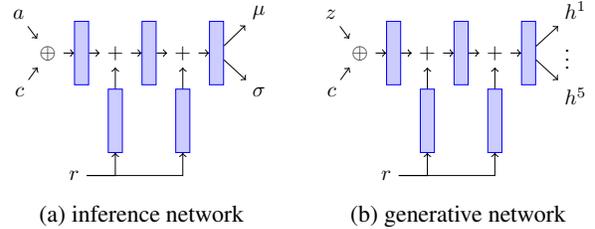
\begin{figure}[t]
  \centering
  \begin{subfigure}[b]{0.49\columnwidth}
    \centering
    \scalebox{0.8}{%
    \begin{tikzpicture}[
      layer/.style={rectangle,draw=blue,fill=blue!20,minimum height=3em},
    ]
      \node[layer] (h1) {};
      \node (a) [above left=-0.25em and 2em of h1] {$a$};
      \node (concat) [left=0.5em of h1] {$\oplus$};
      \node (c) [below left=-0.25em and 2em of h1] {$c$};
      \node (m1) [right=0.5em of h1] {$+$};
      \node[layer] (i1) [below=1em of m1] {};
      \node (r) [below left=0.5em and 1em of i1] {$r$};
      \node[layer] (h2) [right=0.5em of m1] {};
      \node (m2) [right=0.5em of h2] {$+$};
      \node[layer] (i2) [below=1em of m2] {};
      \node[layer] (h3) [right=0.5em of m2] {};
      \node (mu) [above right=-0.25em and 1em of h3] {$\mu$};
      \node (sig) [below right=-0.25em and 1em of h3] {$\sigma$};

      \draw[->] (a) -- (concat);
      \draw[->] (c) -- (concat);
      \draw[->] (concat) -- (h1);
      \draw[->] (r) -| (i1);
      \draw[->] (r) -| (i2);
      \draw[->] (h1) -- (m1);
      \draw[->] (m1) -- (h2);
      \draw[->] (i1) -- (m1);
      \draw[->] (h2) -- (m2);
      \draw[->] (m2) -- (h3);
      \draw[->] (i2) -- (m2);
      \draw[->] (h3) -- (mu);
      \draw[->] (h3) -- (sig);
    \end{tikzpicture}
    }
    \caption{inference network}\label{fig:cvae-enc}
  \end{subfigure}
  \hfill
  \begin{subfigure}[b]{0.49\columnwidth}
    \centering
    \scalebox{0.8}{%
    \begin{tikzpicture}[
      layer/.style={rectangle,draw=blue,fill=blue!20,minimum height=3em},
    ]
      \node[layer] (h1) {};
      \node (z) [above left=-0.25em and 2em of h1] {$z$};
      \node (concat) [left=0.5em of h1] {$\oplus$};
      \node (c) [below left=-0.25em and 2em of h1] {$c$};
      \node (m1) [right=0.5em of h1] {$+$};
      \node[layer] (i1) [below=1em of m1] {};
      \node (r) [below left=0.5em and 1em of i1] {$r$};
      \node[layer] (h2) [right=0.5em of m1] {};
      \node (m2) [right=0.5em of h2] {$+$};
      \node[layer] (i2) [below=1em of m2] {};
      \node[layer] (h3) [right=0.5em of m2] {};
      \node (p1) [above right=-0.25em and 1em of h3] {$h^1$};
      \node (pdots) [right=1em of h3] {$\vdots$};
      \node (p5) [below right=-0.25em and 1em of h3] {$h^5$};

      \draw[->] (z) -- (concat);
      \draw[->] (c) -- (concat);
      \draw[->] (concat) -- (h1);
      \draw[->] (r) -| (i1);
      \draw[->] (r) -| (i2);
      \draw[->] (h1) -- (m1);
      \draw[->] (m1) -- (h2);
      \draw[->] (i1) -- (m1);
      \draw[->] (h2) -- (m2);
      \draw[->] (m2) -- (h3);
      \draw[->] (i2) -- (m2);
      \draw[->] (h3) -- (p1);
      \draw[->] (h3) -- (p5);
    \end{tikzpicture}
    }
    \caption{generative network}\label{fig:cvae-dec}
  \end{subfigure}
  \caption{CVAE with multiplicative injection layers for $r$. $\oplus$ denotes concatenation. During training, action $a$, condition $c$ and reward $r$ are fed to the inference network to get parameters of the variational posterior, $\mu,\sigma$; $z$ is then sampled from $\mathcal{N(\mu,\sigma)}$ and, together with $c$ and $r$, go into the generative network to produce the action distribution . At inference time, $z$ is sampled from the prior $p_0$ instead.}\label{fig:cvae}
\end{figure}

\paragraph{Reward conditioned policies via CVAEs}
We implement the reward-conditioned policies $\pi_\theta(a \mid c, r)$ using CVAEs \citep{kingma2014semisupervisedlearning} with the multiplicative injection layers \citep{oord2016conditionalimage,devries2017modulatingearly,perez2017filmvisual,kumar2019reward} for reward conditioning.\footnote{A naive alternative to the multiplicative injection layers is to concatenate $r$ together with $a$ and $c$. However, we find that in this way $r$ can be ignored in later layers in the neural network, making the learned reward-conditioned policy insensitive to $r$. Similarly findings have been also reported by \citet{kumar2019reward}.}
The CVAE consists an inference network $q_\theta(z \mid a, c, r)$ and a generative network $p_\theta(a \mid c, r, z)$.
The inference network $q_\theta(z \mid a, c, r) := \mathcal{N}(z; \mu_\theta(a, c, r), \sigma_\theta(a, c, r))$ is a parametric normal where $\mu_\theta$ and $\sigma_\theta$ share a few multiplicative injection layers except the last layer that outputs the mean and standard deviation respectively.
The generative network $p_\theta(a \mid c, r, z) := \mathrm{Cat}(a_1; h^1_\theta(c, r, z)) \times \cdots \times \mathrm{Cat}(a_5; h^5_\theta(c, r, z))$ is a product of categorical distributions where $h^1_\theta, \dots, h^5_\theta$ share the layers with multiplicative injection except the last layer that outputs the probability vector (with softmax activation) for each action variable $a_i$.
A schematics of the CVAE is shown in \figref{fig:cvae}.
The CVAE is trained by maximizing the standard ELBO with a standard normal $p_0(z) = \mathcal{N}(0, 1)$ as the prior. 
We also down-weight the Kullback-Leibler (KL) divergence term by a weight $\beta=0.5$, i.e.~annealing, to mitigate component collapse during training \citep{sonderby2016laddervariational}.
When new observations are returned by the delay buffer, we retrain the CAVE from scratch for a fixed number of total steps with all the data observed so far.
After training, the reward conditioned policy can be approximated by Monte Carlo estimation using $N_z$ samples of $z$ from the prior $p_0$ as 
$$
\begin{aligned}
\pi_\theta(a \mid c, r) 
&= \E_{z \sim p_0} [p_\theta(a \mid c, r, z)] \\
&\approx {1 \over N_z} \sum_{i=1}^{N_z} p_\theta(a \mid c, r, z^i)
\end{aligned},
$$
where $z^i \sim p_0(z)$; in practice, we find that only 3--5 samples are enough for stable and consistent policies.

\begin{figure}[t]
  \centering
  \includegraphics[width=0.32\textwidth]{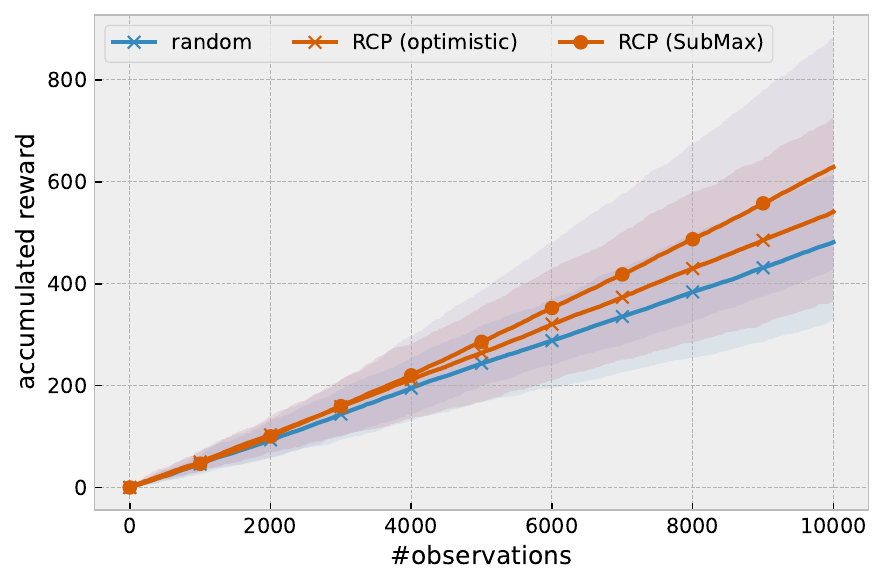}
  \caption{Accumulated reward for simulations w/ $\E[\bar{R}] = 0.0483$.}
  \label{fig:exp-3}
\end{figure}
The accumulated reward plot is given in \figref{fig:exp-3}, where we also provide the expected reward under a uniformly random policy, $\E[\bar{R}] = 0.0483$, to give a sense of the high reward sparsity.
As it can be seen, RCPs with both strategies can be successfully applied to this MAB setting with an action space of cardinality $32768$, yielding better than chance performance.
Between the two strategies, the SubMax strategy still consistently works better than the optimistic strategy.
This indicates its potential effectiveness in design tasks where multiple discrete design choices need to be made jointly.

\section{Discussions, Limitations and Future Work}\label{sec:discussion}

\paragraph{Relationships to existing methods}
Many other methods have the notion of measuring the contrast between data with positive and negative rewards.
In gradient bandits, the gradient has a similar contrastive form \citep{sutton2018reinforcementlearning}.
\citet{kumar2019reward} studies how the advantages, defined as the differences between estimated state-action value and estimated state value, can be used for conditioning.
It can be seen as contrastive against a base reward estimated by an extra reward model.
More recently, direct policy optimization \citep{rafailov2023directpreference} also involves a difference term using positive and negative samples in its loss/gradient computation.

\paragraph{Decoupled learning and inference polices}
An important take-away from our study is that by taking use of the fact that the policies at learning time and inference time can be different in RCPs, there is room to further optimize the inference-time policies.
More research could be done in this direction of decoupled policies in learning and inference to derive better MAB and RL algorithms.

\paragraph{Theoretical analysis}
Although the effectiveness of our proposed strategies to construct policies in RCPs have been empirically validated through simulation, the theoretical foundation is still missing.
In particular, it would be interesting to derive the convergence bound for each strategy and compare the rates against each other and further understand the effectiveness of the SubMax strategy.

\paragraph{Extending to real-valued actions}
Our presentation has mostly focused on developing strategies for discrete actions.
Real-valued actions are challenging because it is hard to construct a valid and tractable policy through generalized marginalization.
For optimized policies, the constraints in \eqref{eq:bounds} would be infinitely many.
For SubMax polices, it is generally hard to sample from the resulting policy due to its unknown normalization constant, which would require computationally expansive sampling methods like Markov chain Monte Carlo algorithms.

\paragraph{General ways to handle beyond binary rewards}
\Secref{sec:other-reward-types} presents ways to handle general discrete rewards and real-valued rewards, but they rely on the simplification of using only two specific reward conditions (low and high rewards), similar to the binary case.
However, a general extension, e.g.~to real-valued rewards, is non-trivial as in \eqref{eq:maximizing-expected-reward} it would require optimizing/learning a continuous weight function with the constraint of having a resulting integral of $1$.\looseness=-1

\paragraph{Continual learning of deep generative models}
In \secref{sec:sim-design}, we retrain the CVAE from scratch for each round, mostly for its simplicity to avoid catastrophic forgetting.
However, this wastes computation and methods like variational continual learning \citep{kingma2014semisupervisedlearning} can be potentially used to avoid training from scratch, which is more practical if the model is large.
In general, how to continually train neural network based policies with new data without re-initializing the networks is an important research question.

\section{Conclusions}

We have presented a novel approach to improve the convergence of RCPs using generalized marginalization and studied several concrete strategies in MAB with discrete action spaces.
Practically, we have identified a simple-to-implement strategy, the SubMax policy, effective in a diverse set of simulation settings.
We hope our approach can make RCPs more competitive in various MAB settings and enable easy development of polices implemented by deep generative models to solve bandit problems.

\newpage




\bibliographystyle{icml2024}
\bibliography{refs}

\newpage
\appendix
\onecolumn

\section{Results for all 27 configurations in \secref{sec:sim-challenging}}\label{app:exp-challenging}
\begin{figure}[t]
  \centering
  \vspace{-5em}
  \begin{small}
  \begin{subfigure}[b]{0.24\textwidth}
      \centering
      \includegraphics[width=\textwidth]{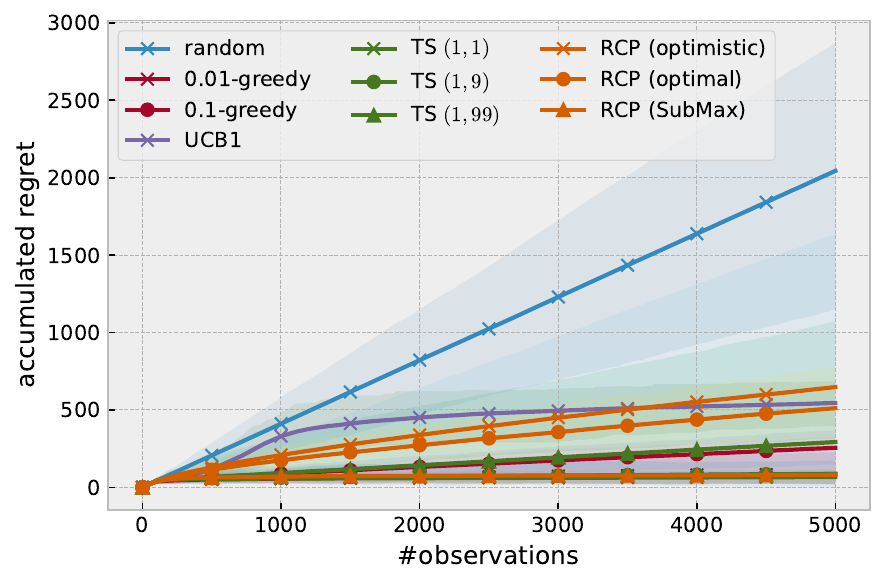}
      \caption{$K=10, (\alpha, \beta)=(1, 1), N_b=100$}
      \label{fig:exp-1-K=10-Nb=100-ab=uniform}
  \end{subfigure}
  \hfill
  \begin{subfigure}[b]{0.24\textwidth}
      \centering
      \includegraphics[width=\textwidth]{./figs/exp-1/K=100-Nb=100-ab=uniform.pdf}
      \caption{$K=100, (\alpha, \beta)=(1, 1), N_b=100$}
      \label{fig:exp-1-K=100-Nb=100-ab=uniform}
  \end{subfigure}
  \hfill
  \begin{subfigure}[b]{0.24\textwidth}
      \centering
      \includegraphics[width=\textwidth]{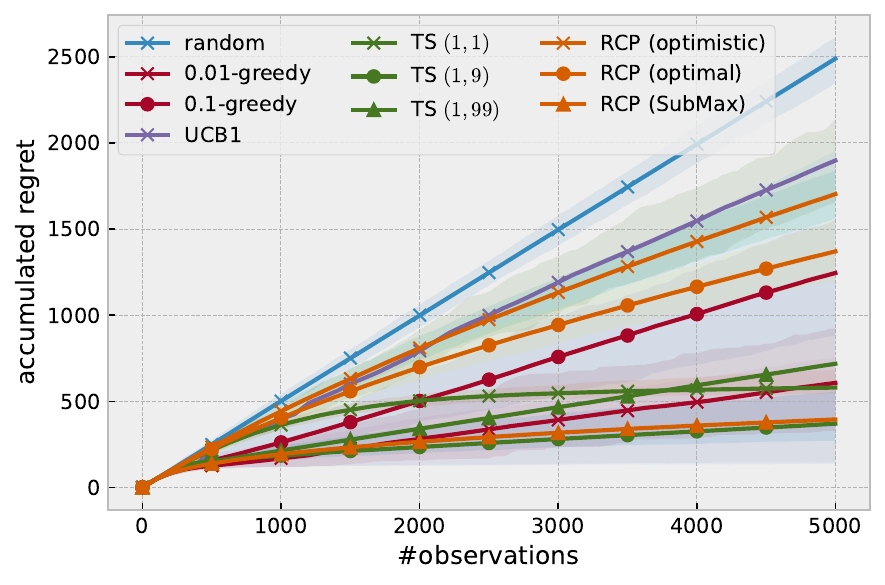}
      \caption{$K=500, (\alpha, \beta)=(1, 1), N_b=100$}
      \label{fig:exp-1-K=500-Nb=100-ab=uniform}
  \end{subfigure}
  \hfill
  \begin{subfigure}[b]{0.24\textwidth}
    \centering
    \includegraphics[width=\textwidth]{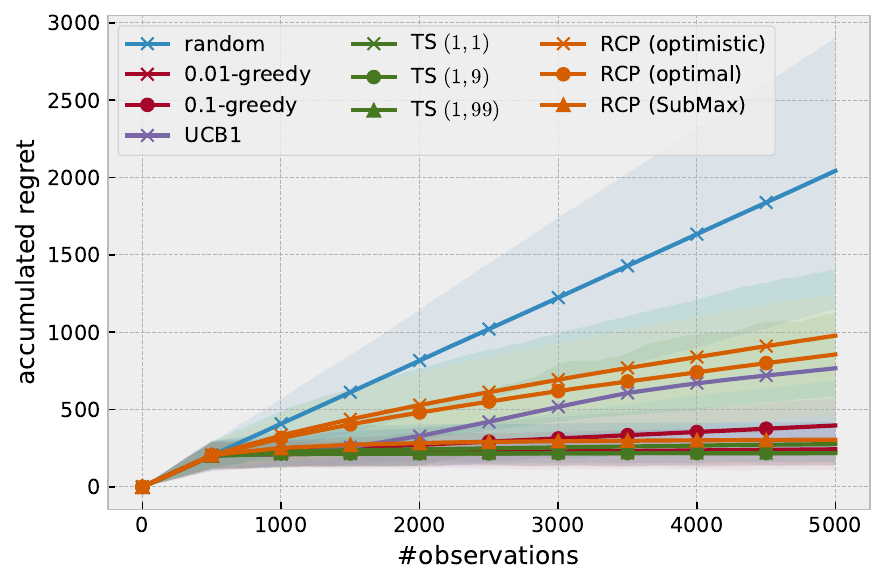}
    \caption{$K=10, (\alpha, \beta)=(1, 1), N_b=500$}
    \label{fig:exp-1-K=10-Nb=500-ab=uniform}
\end{subfigure}
\hfill
\begin{subfigure}[b]{0.24\textwidth}
    \centering
    \includegraphics[width=\textwidth]{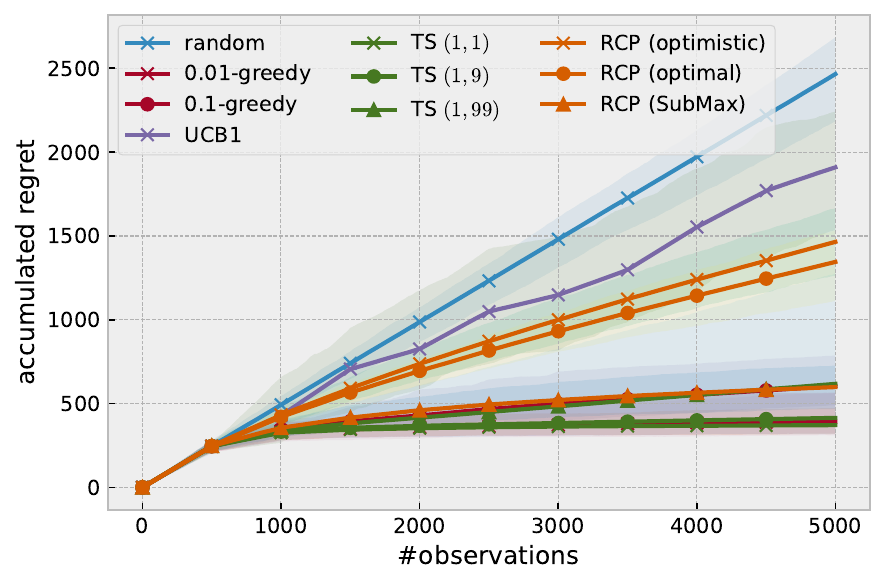}
    \caption{$K=100, (\alpha, \beta)=(1, 1), N_b=500$}
    \label{fig:exp-1-K=100-Nb=500-ab=uniform}
\end{subfigure}
\hfill
\begin{subfigure}[b]{0.24\textwidth}
    \centering
    \includegraphics[width=\textwidth]{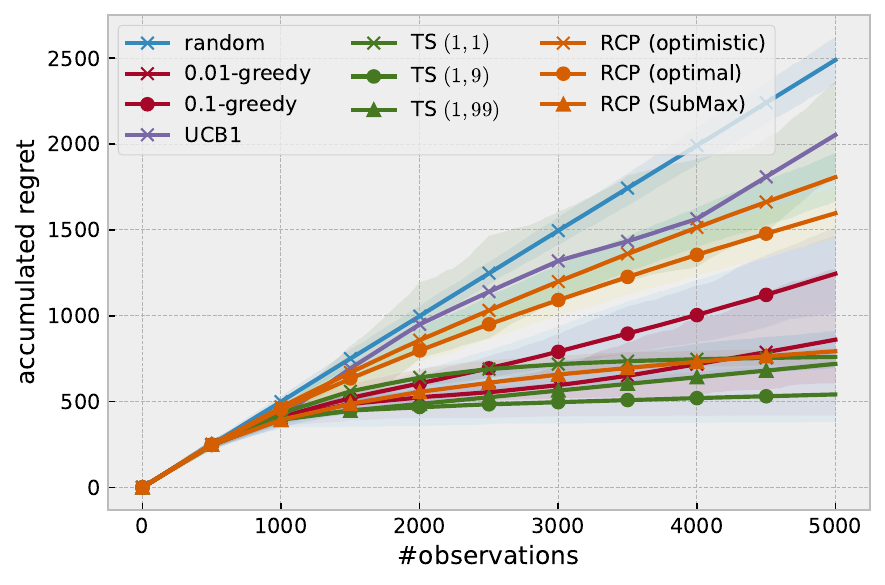}
    \caption{$K=500, (\alpha, \beta)=(1, 1), N_b=500$}
    \label{fig:exp-1-K=500-Nb=500-ab=uniform}
\end{subfigure}
\hfill
\begin{subfigure}[b]{0.24\textwidth}
  \centering
  \includegraphics[width=\textwidth]{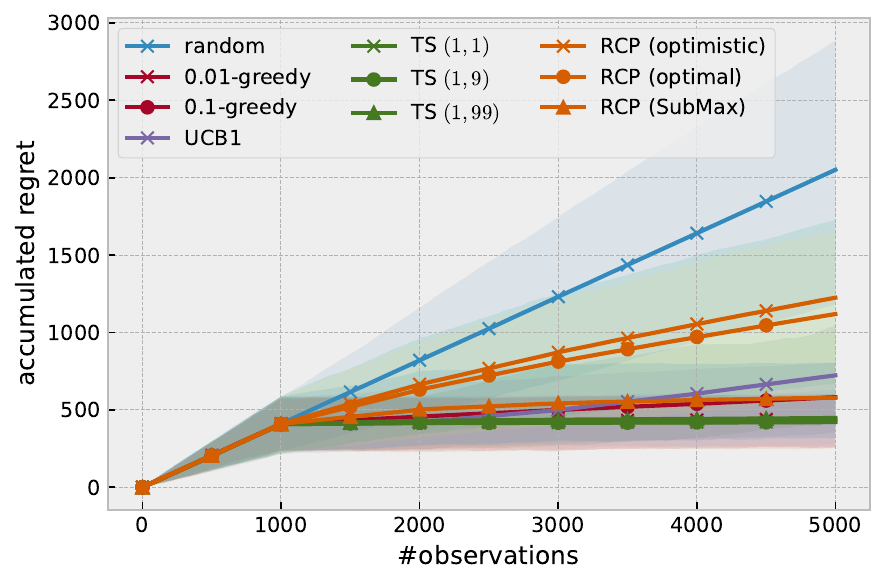}
  \caption{$K=10, (\alpha, \beta)=(1, 1), N_b=1000$}
  \label{fig:exp-1-K=10-Nb=1000-ab=uniform}
\end{subfigure}
\hfill
\begin{subfigure}[b]{0.24\textwidth}
  \centering
  \includegraphics[width=\textwidth]{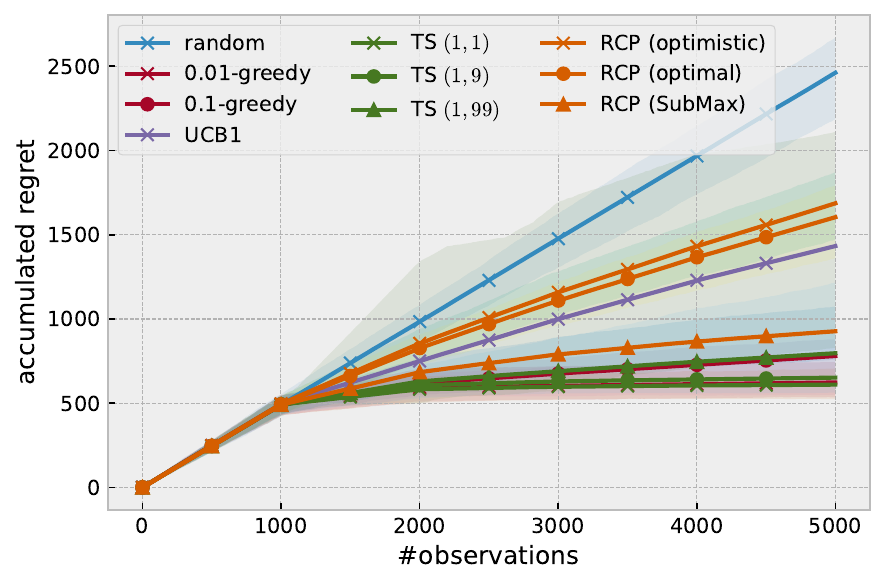}
  \caption{$K=100, (\alpha, \beta)=(1, 1), N_b=1000$}
  \label{fig:exp-1-K=100-Nb=1000-ab=uniform}
\end{subfigure}
\hfill
\begin{subfigure}[b]{0.24\textwidth}
  \centering
  \includegraphics[width=\textwidth]{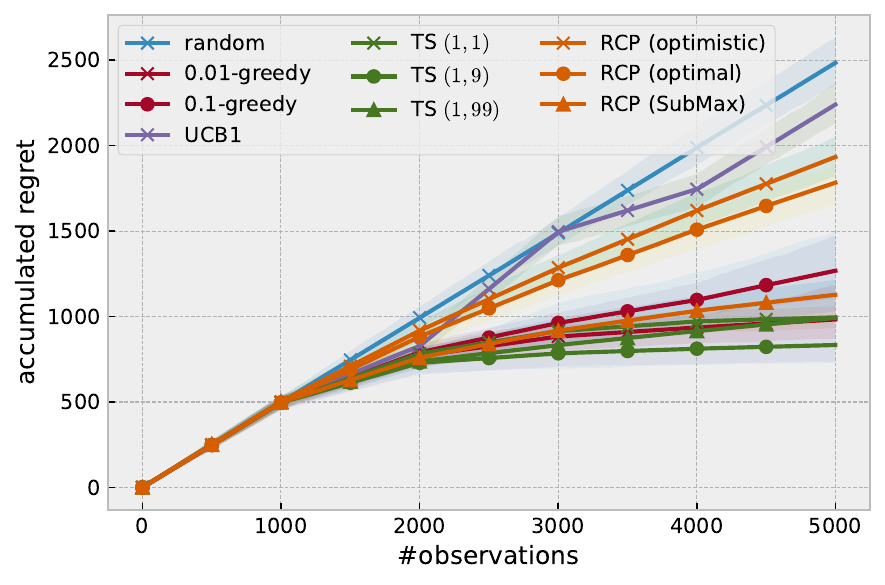}
  \caption{$K=500, (\alpha, \beta)=(1, 1), N_b=1000$}
  \label{fig:exp-1-K=500-Nb=1000-ab=uniform}
\end{subfigure}
\hfill
\begin{subfigure}[b]{0.24\textwidth}
      \centering
      \includegraphics[width=\textwidth]{./figs/exp-1/K=10-Nb=100-ab=sparse.pdf}
      \caption{$K=10, (\alpha, \beta)=(1, 9), N_b=100$}
      \label{fig:exp-1-K=10-Nb=100-ab=sparse}
  \end{subfigure}
  \hfill
  \begin{subfigure}[b]{0.24\textwidth}
      \centering
      \includegraphics[width=\textwidth]{./figs/exp-1/K=100-Nb=100-ab=sparse.pdf}
      \caption{$K=100, (\alpha, \beta)=(1, 9), N_b=100$}
      \label{fig:exp-1-K=100-Nb=100-ab=sparse}
  \end{subfigure}
  \hfill
  \begin{subfigure}[b]{0.24\textwidth}
      \centering
      \includegraphics[width=\textwidth]{./figs/exp-1/K=500-Nb=100-ab=sparse.pdf}
      \caption{$K=500, (\alpha, \beta)=(1, 9), N_b=100$}
      \label{fig:exp-1-K=500-Nb=100-ab=sparse}
  \end{subfigure}
  
  \begin{subfigure}[b]{0.24\textwidth}
    \centering
    \includegraphics[width=\textwidth]{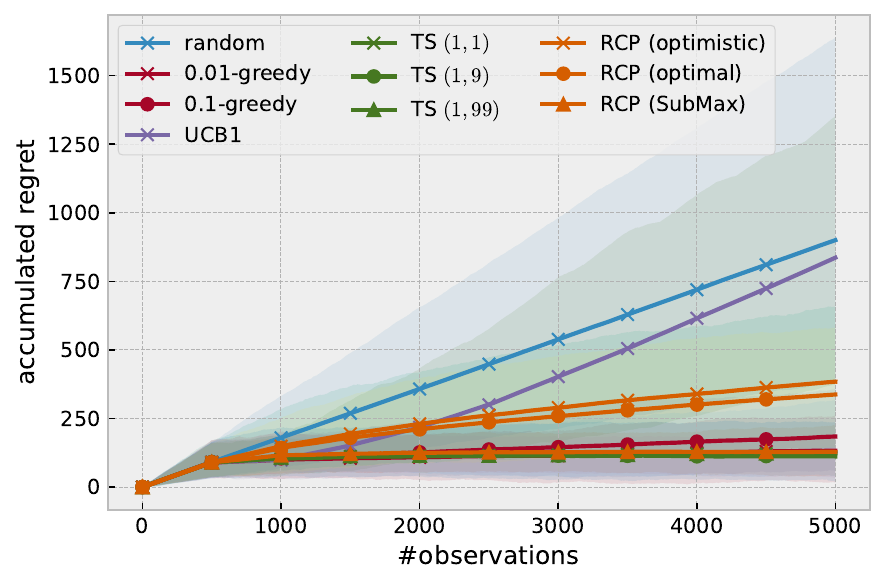}
    \caption{$K=10, (\alpha, \beta)=(1, 9), N_b=500$}
    \label{fig:exp-1-K=10-Nb=500-ab=sparse}
\end{subfigure}
\hfill
\begin{subfigure}[b]{0.24\textwidth}
    \centering
    \includegraphics[width=\textwidth]{./figs/exp-1/K=100-Nb=500-ab=sparse.pdf}
    \caption{$K=100, (\alpha, \beta)=(1, 9), N_b=500$}
    \label{fig:exp-1-K=100-Nb=500-ab=sparse}
\end{subfigure}
\hfill
\begin{subfigure}[b]{0.24\textwidth}
    \centering
    \includegraphics[width=\textwidth]{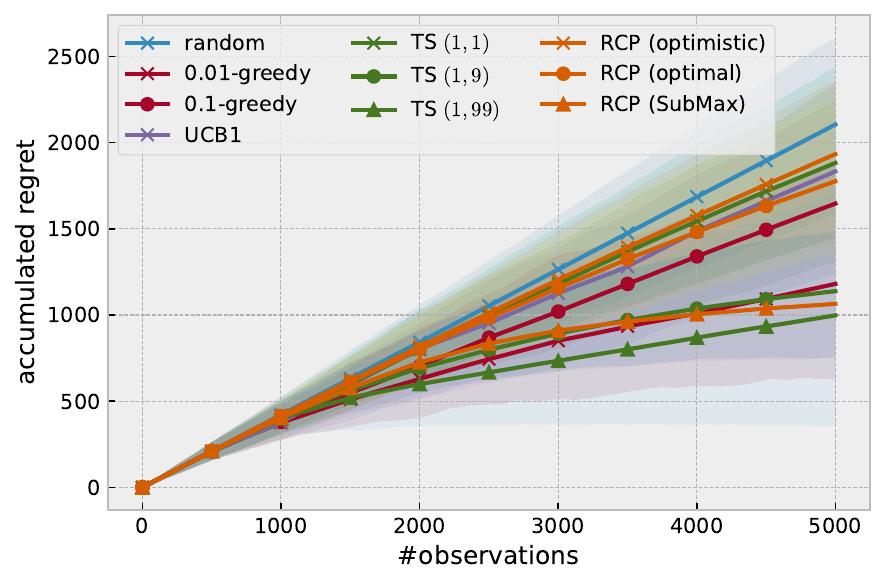}
    \caption{$K=500, (\alpha, \beta)=(1, 9), N_b=500$}
    \label{fig:exp-1-K=500-Nb=500-ab=sparse}
\end{subfigure}
\hfill
\begin{subfigure}[b]{0.24\textwidth}
  \centering
  \includegraphics[width=\textwidth]{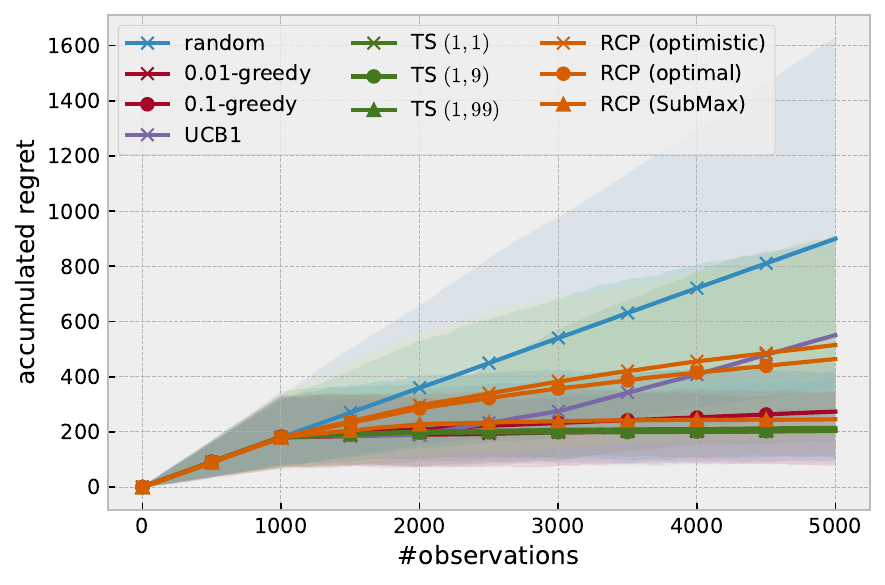}
  \caption{$K=10, (\alpha, \beta)=(1, 9), N_b=1000$}
  \label{fig:exp-1-K=10-Nb=1000-ab=sparse}
\end{subfigure}
\hfill
\begin{subfigure}[b]{0.24\textwidth}
  \centering
  \includegraphics[width=\textwidth]{./figs/exp-1/K=100-Nb=1000-ab=sparse.pdf}
  \caption{$K=100, (\alpha, \beta)=(1, 9), N_b=1000$}
  \label{fig:exp-1-K=100-Nb=1000-ab=sparse}
\end{subfigure}
\hfill
\begin{subfigure}[b]{0.24\textwidth}
  \centering
  \includegraphics[width=\textwidth]{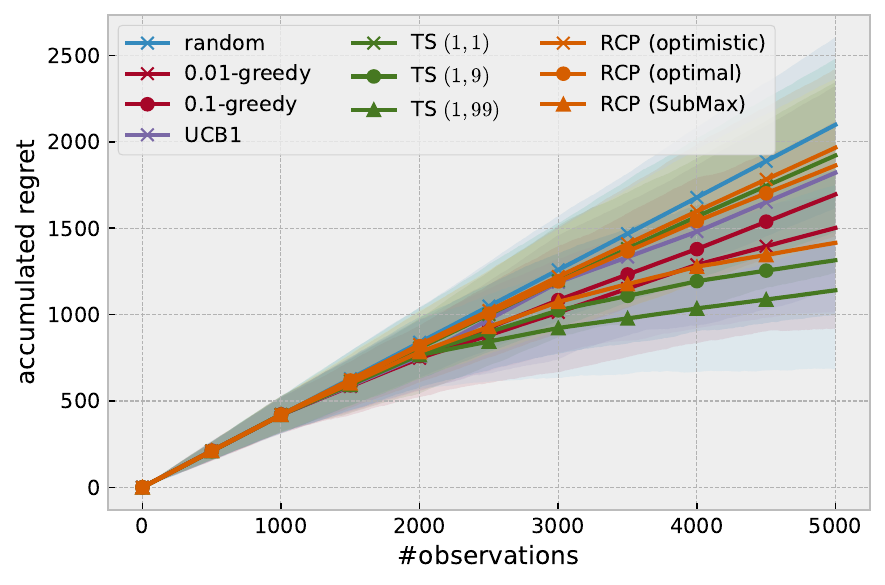}
  \caption{$K=500, (\alpha, \beta)=(1, 9), N_b=1000$}
  \label{fig:exp-1-K=500-Nb=1000-ab=sparse}
\end{subfigure}
\hfill
\begin{subfigure}[b]{0.24\textwidth}
  \centering
  \includegraphics[width=\textwidth]{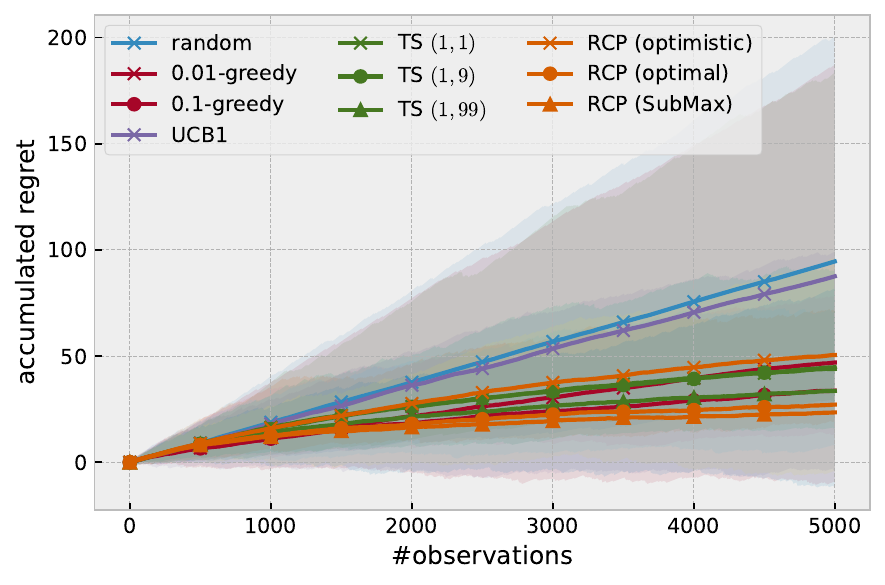}
  \caption{$K=10, (\alpha, \beta)=(1, 99), N_b=100$}
  \label{fig:exp-1-K=10-Nb=100-ab=xsparse}
\end{subfigure}
\hfill
\begin{subfigure}[b]{0.24\textwidth}
  \centering
  \includegraphics[width=\textwidth]{./figs/exp-1/K=100-Nb=100-ab=xsparse.pdf}
  \caption{$K=100, (\alpha, \beta)=(1, 99), N_b=100$}
  \label{fig:exp-1-K=100-Nb=100-ab=xsparse}
\end{subfigure}
\hfill
\begin{subfigure}[b]{0.24\textwidth}
  \centering
  \includegraphics[width=\textwidth]{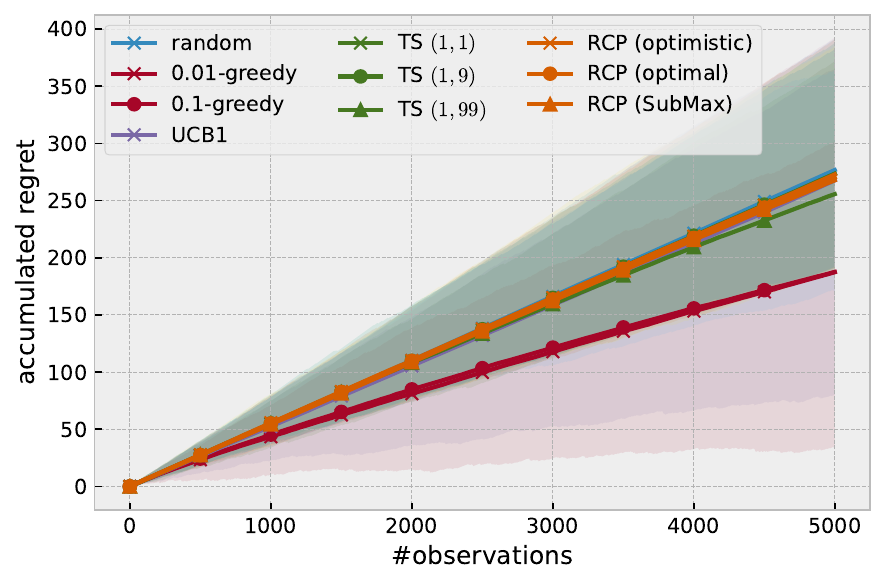}
  \caption{$K=500, (\alpha, \beta)=(1, 99), N_b=100$}
  \label{fig:exp-1-K=500-Nb=100-ab=xsparse}
\end{subfigure}
\hfill
\begin{subfigure}[b]{0.24\textwidth}
\centering
\includegraphics[width=\textwidth]{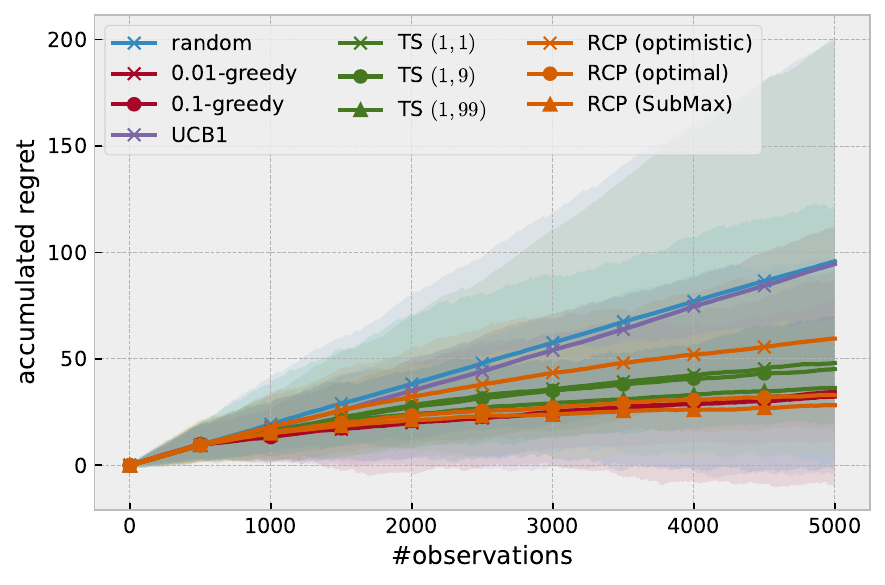}
\caption{$K=10, (\alpha, \beta)=(1, 99), N_b=500$}
\label{fig:exp-1-K=10-Nb=500-ab=xsparse}
\end{subfigure}
\hfill
\begin{subfigure}[b]{0.24\textwidth}
\centering
\includegraphics[width=\textwidth]{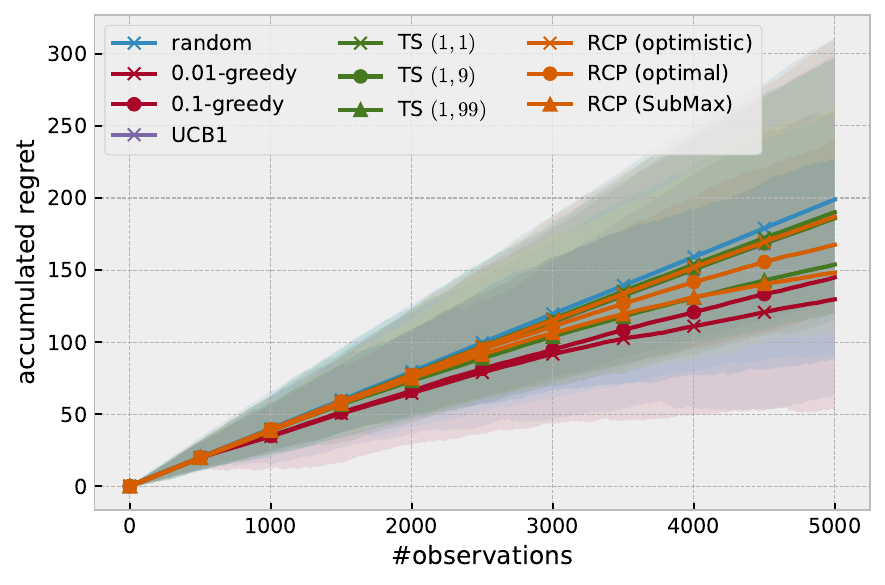}
\caption{$K=100, (\alpha, \beta)=(1, 99), N_b=500$}
\label{fig:exp-1-K=100-Nb=500-ab=xsparse}
\end{subfigure}
\hfill
\begin{subfigure}[b]{0.24\textwidth}
\centering
\includegraphics[width=\textwidth]{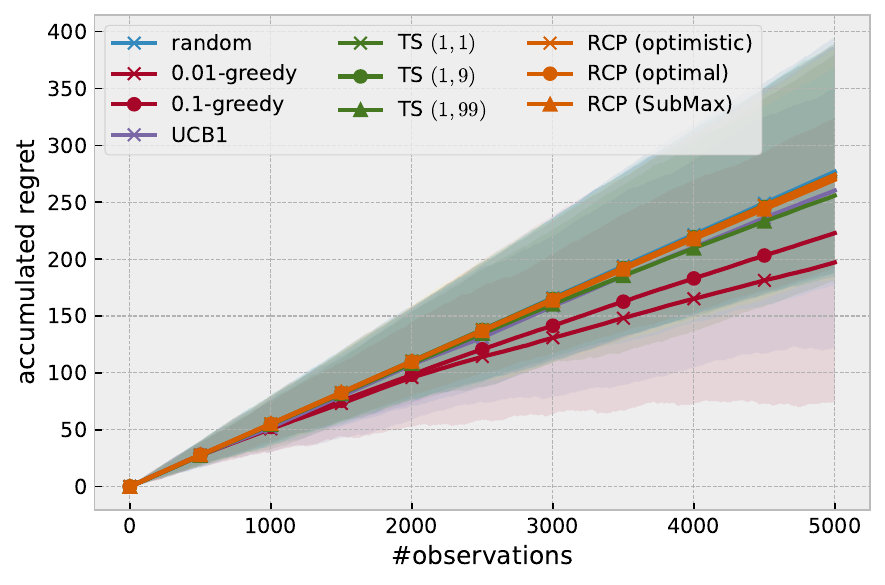}
\caption{$K=500, (\alpha, \beta)=(1, 99), N_b=500$}
\label{fig:exp-1-K=500-Nb=500-ab=xsparse}
\end{subfigure}
\hfill
\begin{subfigure}[b]{0.24\textwidth}
\centering
\includegraphics[width=\textwidth]{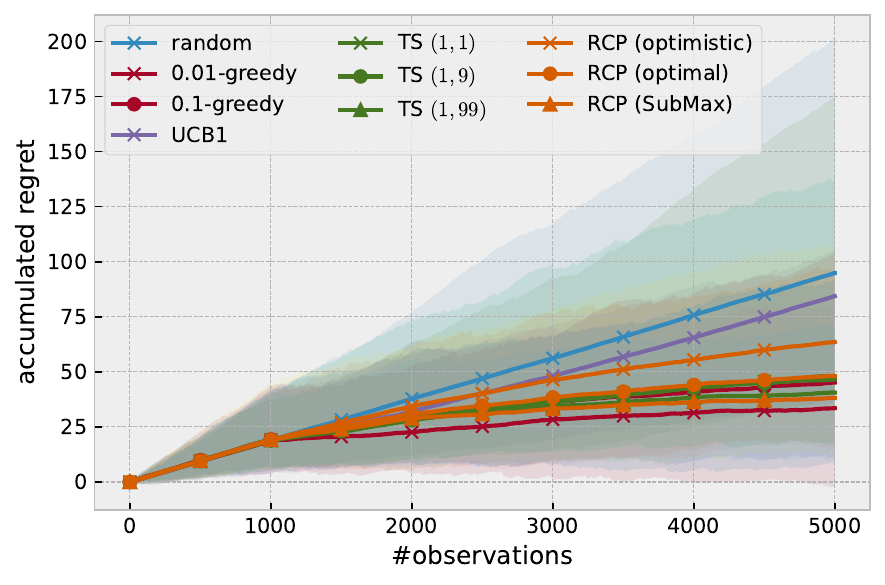}
\caption{$K=10, (\alpha, \beta)=(1, 99), N_b=1000$}
\label{fig:exp-1-K=10-Nb=1000-ab=xsparse}
\end{subfigure}
\hfill
\begin{subfigure}[b]{0.24\textwidth}
\centering
\includegraphics[width=\textwidth]{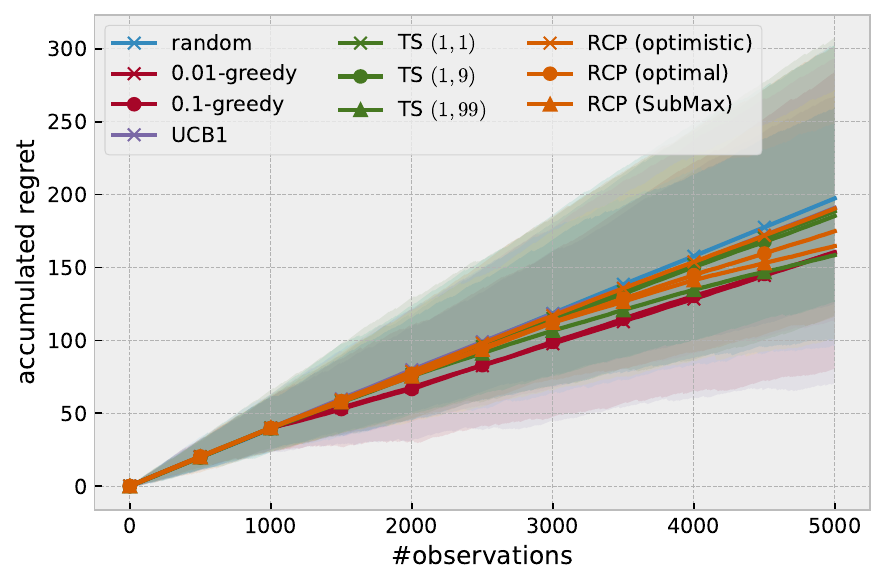}
\caption{$K=100, (\alpha, \beta)=(1, 99), N_b=1000$}
\label{fig:exp-1-K=100-Nb=1000-ab=xsparse}
\end{subfigure}
\hfill
\begin{subfigure}[b]{0.24\textwidth}
\centering
\includegraphics[width=\textwidth]{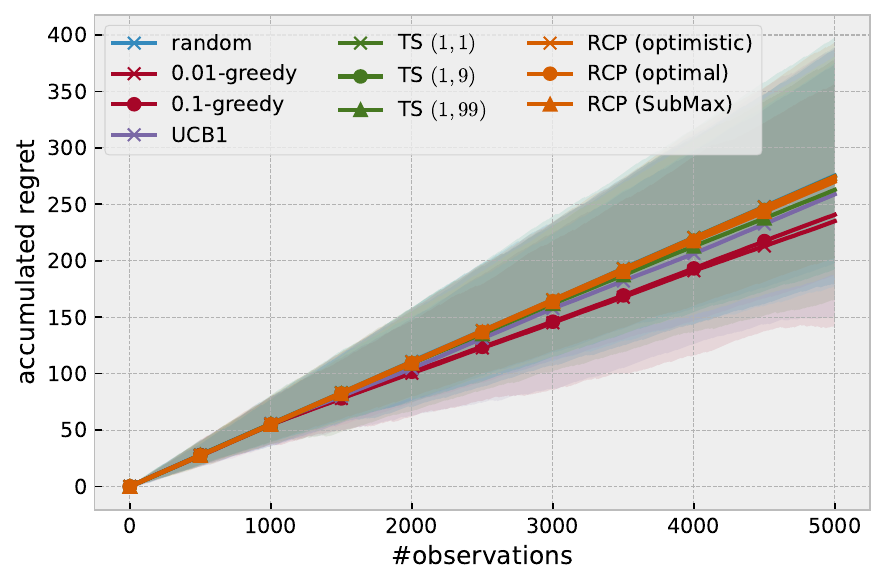}
{\small (1) $K=500, (\alpha, \beta)=(1, 99), N_b=1000$}
\label{fig:exp-1-K=500-Nb=1000-ab=xsparse}
\end{subfigure}
\end{small}
  \caption{Accumulated regret with varying $K$, $(\alpha, \beta)$ and $N_b$ (full results for \secref{sec:sim-challenging}).}
  \label{fig:exp-1-all}
\end{figure}

\end{document}